\documentclass[letterpaper,journal]{IEEEtran}
\usepackage{amsmath,amsfonts}
\usepackage{algorithmic}
\usepackage{algorithm}
\usepackage{array}
\usepackage[caption=false,font=normalsize,labelfont=sf,textfont=sf]{subfig}
\usepackage{textcomp}
\usepackage{stfloats}
\usepackage{url}
\usepackage{verbatim}
\usepackage{graphicx}
\usepackage{cite}
\usepackage{xcolor}
\usepackage{makecell}
\usepackage[colorlinks=true,linkcolor=blue,citecolor=blue]{hyperref}
\usepackage{pifont}
\usepackage{multirow}
\usepackage{booktabs}

\begin{document}

\title{4D-MoDe: Towards Editable and Scalable Volumetric Streaming via Motion-Decoupled 4D Gaussian Compression}

\author{
Houqiang~Zhong,
~Zihan~Zheng,
~Qiang~Hu,
~\textit{Member, IEEE},
~Yuan~Tian,
~Ning~Cao,
~Lan~Xu,
~Xiaoyun~Zhang,

~\textit{Member, IEEE},
~Zhengxue~Cheng,
~Li~Song,~\textit{Senior Member, IEEE},
~Wenjun~Zhang,~\textit{Fellow, IEEE}
\thanks{
Houqiang Zhong, Zihan Zheng, Qiang Hu, Xiaoyun Zhang, Zhengxue Cheng, Li Song and Wenjun Zhang are with the Shanghai Jiao Tong University, Shanghai, 200240, China. (email:{zhonghouqiang, 1364406834, qiang.hu, xiaoyun.zhang, zxcheng, 
song\_li, zhangwenjun}@sjtu.edu.cn)

Yuan Tian is with the Shanghai Artificial Intelligence Laboratory, Shanghai, 200232, China. (email: tianyuan168326@outlook.com)

Ning Cao is with the E-surfing Vision Technology Co., Ltd, Hangzhou, 311100, China. (email:caoning.sh@chinatelecom.cn)

Lan Xu is with the ShanghaiTech University, Shanghai, 201210, China. (email:xulan1@shanghaitech.edu.cn)
\textit{(Corresponding author: Qiang Hu).}
}
}

\markboth{
IEEE Transactions On Circuits and Systems For Video Technology,~Vol.~14, No.~8, May~2025}%
{Zhong \MakeLowercase{\textit{et al.}}: 4D-MoDe: Towards Editable and Scalable Volumetric Streaming via
Motion-Decoupled 4D Gaussian Compression}

\maketitle

\begin{abstract}

Volumetric video has emerged as a key medium for immersive telepresence and augmented/virtual reality, enabling six-degrees-of-freedom (6DoF) navigation and realistic spatial interactions. However, delivering high-quality dynamic volumetric content at scale remains challenging due to massive data volume, complex motion, and limited editability of existing representations. In this paper, we present 4D-MoDe, a motion-decoupled 4D Gaussian compression framework designed for scalable and editable volumetric video streaming. Our method introduces a layered representation that explicitly separates static backgrounds from dynamic foregrounds using a lookahead-based motion decomposition strategy, significantly reducing temporal redundancy and enabling selective background/foreground streaming. To capture continuous motion trajectories, we employ a multi-resolution motion estimation grid and a lightweight shared MLP, complemented by a dynamic Gaussian compensation mechanism to model emergent content. An adaptive grouping scheme dynamically inserts background keyframes to balance temporal consistency and compression efficiency. Furthermore, an entropy-aware training pipeline jointly optimizes the motion fields and Gaussian parameters under a rate-distortion (RD) objective, while employing range-based and KD-tree compression to minimize storage overhead. Extensive experiments on multiple datasets demonstrate that 4D-MoDe consistently achieves competitive reconstruction quality with an order of magnitude lower storage cost (e.g., as low as \textbf{11.4} KB/frame) compared to state-of-the-art methods, while supporting practical applications such as background replacement and foreground-only streaming.
\end{abstract}


\section{Introduction}
Volumetric video is rapidly emerging as a pivotal medium for immersive experiences in telepresence, augmented reality (AR), and virtual reality (VR). Unlike conventional 2D or multiview formats, it captures the spatiotemporal 3D geometry and appearance of dynamic scenes, enabling realistic spatial interactions and six degrees of freedom (6DoF) navigation, crucial for next-generation immersive media applications.
However, scaling delivery and rendering of high-quality volumetric video remains a major challenge, especially for dynamic scenes with significant motion, complex backgrounds, and extended durations. The massive data volume, intensive computational demands, and limited editability of existing representations severely constrain the scalability, interactivity, and adaptability of current solutions, particularly under bandwidth constraints.

Traditional volumetric video reconstruction methods predominantly rely on geometry-based techniques such as dynamic meshes\cite{HumanMeshRecovery,hsmr}, point clouds~\cite{PointCloudBasedVolumetricVideoCodecs,DynamicPointCloud}, depth maps \cite{fvvdibr} and image-based view interpolation \cite{varfvv}. While these approaches facilitate real-time rendering to some extent, they commonly suffer from limited reconstruction accuracy and visual artifacts, particularly in 
occlusions, textureless surfaces, or intricate details like hair and foliage. Neural Radiance Fields (NeRF) \cite{nerf} offered a breakthrough in photorealism, mitigating many reconstruction artifacts common in traditional geometric methods. 
Various approaches\cite{StreamRF,DNeRF,neuralsceneflowfields,SpaceTimeNeuralFields,DynamicNeuralRadianceFields} extend NeRF to dynamic scenes, but they fail to support long sequences and streaming due to large model size. Consequently, several approaches\cite{rerf,EfficientDynamicNeRF,jointrf,VRVVC,hpc,tetrirf,nerfplayer,CompactNeuralVolumetricVideo} attempt to compress the explicit representations of dynamic NeRFs. For example, ReRF \cite{rerf} leverages a grid-based spatio-temporal representation and applies conventional image compression techniques to reduce redundancy. Despite their effectiveness, such methods typically incur significant computational overhead, limiting their practicality in real-time scenarios.

To overcome the efficiency limitations of NeRF-based methods, 3D Gaussian Splatting (3DGS) \cite{3dgs} has recently been proposed, enabling both high-quality rendering and real-time performance in static scenes. To extend 3DGS to dynamic scenarios, existing methods can be broadly categorized into sequence-based \cite{spacetimegs,4dgs,MotionFactorizationGS,Deformable3DGS,scgs} and frame-wise modeling \cite{3dgstream,dynamic3dgs,v3,SplatterAVideo,hicom2024,queen} strategies. Sequence-based methods optimize over entire video sequences to reduce temporal redundancy but require loading multiple frames into memory simultaneously, making them difficult to scale for long sequences or practical streaming applications. In contrast,  frame-wise approaches, such as 3DGStream \cite{3dgstream}, model frame-to-frame transformations by learning a neural transformation cache that captures the inter-frame motion of 3D Gaussians. 
However, these methods still exhibit substantial inter-frame redundancy, limiting their effectiveness in bandwidth-constrained volumetric video transmission. Although a few recent works~\cite{4DGC,hifi4g,dualgs} have explored compression techniques for dynamic 3DGS, they typically lack explicit decomposition of dynamic foregrounds and static backgrounds, leading to suboptimal compression rates and reduced flexibility in downstream tasks such as editing, object insertion/removal, and interactive rendering.

\begin{figure*}[htb] 
    \centering
    \includegraphics[width=\linewidth]{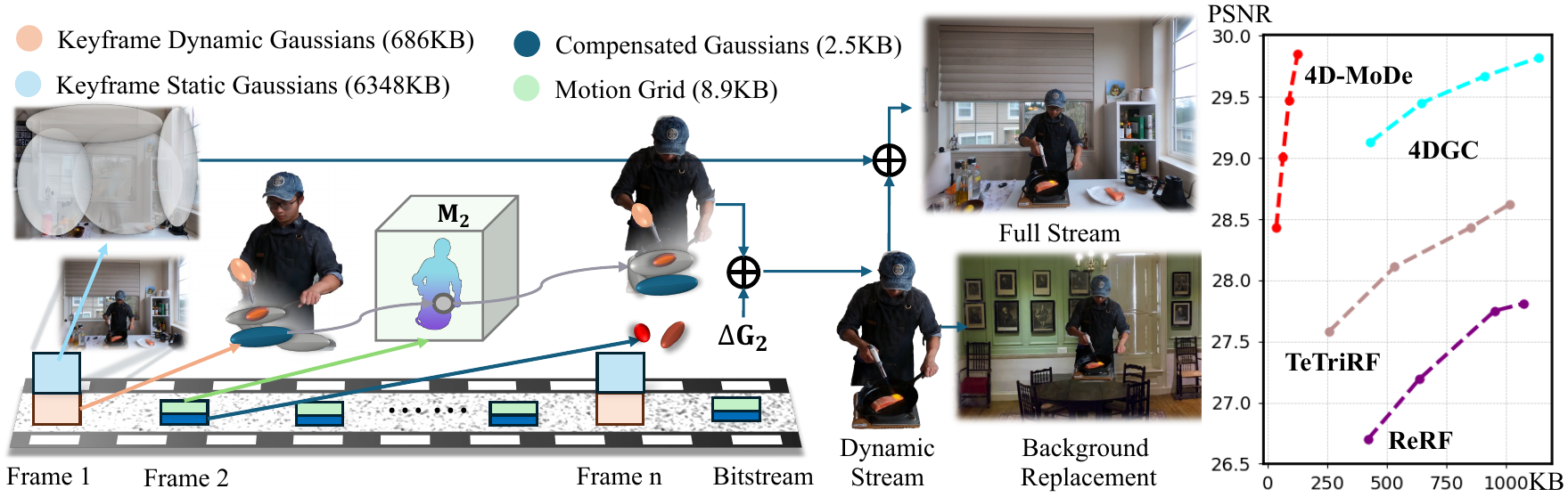}
    \caption{
    \textbf{Left:} With our motion-decoupled 4D representation, 4D-MoDe encodes volumetric video into a scalable, ultra-low bitrate stream (11.4KB per inter-frame), comprising a motion grid (8.9KB) and compensated Gaussians (2.5KB). \textbf{Middle:} This structure enables flexible editing, including background replacement and foreground-only streaming. \textbf{Right:} Our method achieves significantly better rate-distortion performance than existing state-of-the-art approaches (e.g., 4DGC~\cite{4DGC}, TeTriRF~\cite{tetrirf}).
    }
    \label{fig:teaser}
\end{figure*}

To tackle these challenges, we present 4D-MoDe, a motion-decoupled 4D Gaussian compression framework tailored for scalable and editable volumetric video streaming. It achieves high-quality real-time rendering at ultra-low bitrates by disentangling motion and appearance across space and time (see Fig.~\ref{fig:teaser}). The core idea is a layered 4D Gaussian representation that explicitly separates dynamic foreground motion from the static background through a lookahead-based motion decoupling strategy. This structured decomposition facilitates independent modeling and compression of foreground and background components, significantly reducing temporal redundancy and unlocking downstream flexibility such as content editing and interactive control. To model dynamic motion effectively, we introduce a multi-resolution motion estimation grid combined with a lightweight shared MLP, which estimates per-frame translation and rotation parameters for each dynamic Gaussian. This design constructs smooth 4D trajectories for continuous and compact motion modeling, with trilinear interpolation from the grid ensuring temporal coherence and multi-scale fidelity.

To further enhance visual quality in challenging regions, 4D-MoDe incorporates a dynamic Gaussian compensation mechanism, which adaptively inserts new Gaussians in regions with high image-space gradients or abrupt changes, such as newly appearing small objects or localized deformations. For handling background transitions, we propose an adaptive Group-of-Pictures (GOP) scheme that inserts background keyframes only when needed, achieving a balance between background stability and compression efficiency. Additionally, we propose an entropy-optimized training scheme, which jointly tunes motion grids and Gaussian parameters toward a rate-distortion (RD) objective. Our final representation is a spatio-temporally compact set of Gaussians and motion fields.  Redundancy in this representation is minimized using range-based entropy coding for the motion grid, and KD-tree-based\cite{draco} hierarchical clustering for keyframe Gaussians and compensated Gaussians. Extensive experiments on multiple volumetric video datasets demonstrate that 4D-MoDe achieves state-of-the-art RD performance, with an average storage cost of \textbf{11.4} KB per inter-frame.

Our contributions are summarized as follows:
\begin{itemize}
    \item We propose a novel motion-decoupled 4D Gaussian representation that explicitly separates static backgrounds and dynamic foregrounds through a layered structure and lookahead-based decomposition, significantly reducing redundancy while enabling flexible content editing and scalable foreground/background streaming. 
    \item We introduce a multi-resolution motion grid and a lightweight shared MLP to predict continuous Gaussian transformations, combined with a dynamic compensation mechanism to handle complex motions and emerging content at fine granularity.
    \item We present an adaptive grouping mechanism for background modeling, which dynamically adjusts keyframe insertion to effectively manage large-scale variations while preserving temporal consistency and optimizing compression. 
    \item We design a three-stage, entropy-aware training pipeline with range-based and KD-tree compression, achieving state-of-the-art RD performance on real-world datasets and supporting applications such as background replacement and foreground-only streaming.
\end{itemize}

\section{Relatd Work}
\subsection{NeRF-based Volumetric Video Modeling}

Neural Radiance Fields (NeRF)~\cite{nerf} have revolutionized 3D scene representation by leveraging differentiable volume rendering and implicit neural representations. Building on this foundation, numerous methods have enhanced NeRF's compactness and rendering speed for static scenes \cite{mipnerf, mipnerf360, instantngp, nerfinwild}. 
To extend NeRF to dynamic scenes, two primary categories of approaches have emerged. 
Flow-based approaches~\cite{neuralsceneflowfields,DynIBaR} reconstruct 3D features from monocular videos, reducing data collection complexity. However, they often require additional priors or constraints to handle complex real-world motion. Deformation field methods~\cite{NeuralRadianceFlow,Nerfies,DNeRF,nerfplayer,gearnerf} model scene dynamics by warping each frame into a shared canonical space, enabling temporal alignment across frames. Despite their effectiveness in capturing motion, these methods typically suffer from high training costs and slow rendering, limiting their applicability to real-time scenarios. To improve rendering speed and temporal modeling, recent efforts have shifted towards explicit 4D radiance field representations using structured volumetric decompositions, such as voxel grids~\cite{tineuvox,MixVoxels}, multi-plane images\cite{kplanes,hexplane}, and tensor factorizations~\cite{HumanRF,tensor4d,TemporalInterpolationisallYouNeed,DictionaryFields}. These designs support faster rendering and more efficient training by modeling spatial-temporal structures explicitly. However, due to their unified representation of space and time, such methods generally lack motion-background separation and fine-grained control, making them less compatible with streaming applications and scene-level editing.

\subsection{3DGS-based Volumetric Video Modeling}
3DGS\cite{3dgs} has emerged as an efficient and interpretable explicit representation for static scene modeling, combining sparse Gaussian primitives with a rasterization-based rendering pipeline to balance quality and real-time performance. 
Recent works have enhanced training efficiency, rendering fidelity, and scene modeling via improved optimization techniques and architectural innovations~\cite{3DGSR,LiftingbyGaussians,2dgs,GoMAvatar,Triplanegs,GaussianEditor}.
To extend 3DGS to dynamic scenes, existing methods can be broadly categorized into three paradigms. 
The first leverages deformation fields, using MLPs or spatial structures to predict per-frame variations of Gaussian parameters~\cite{Deformable3DGS,GeometryAwareDeformableGS,DreamMesh4D,scgs}. 
The second treats time as an additional dimension, modeling 4D Gaussian primitives that are sliced or projected into 3D during rendering~\cite{spacetimegs,4dgs,MotionFactorizationGS,MotionAware3DGS}. The third category is explicit frame-wise modeling, which updates Gaussian attributes between frames via rigid or non-rigid motion estimation, making it compatible with online and streamable volumetric rendering~\cite{dynamic3dgs,3dgstream,hicom2024,queen,hifi4g}. Despite their advancements, these dynamic 3DGS paradigms commonly entangle static backgrounds and dynamic foregrounds within a unified representation. This often leads to redundancy, inefficient transmission, limited content-level editability and can degrade visual quality. To address this issue, we propose a motion-decoupled 4D Gaussian framework that explicitly separates static and dynamic components. This design improves structural modularity and editability while retaining the rendering efficiency of 3DGS, making it suitable for scalable and interactive volumetric video applications.

\subsection{Volumetric Video Compression}
The effectiveness of volumetric video compression is tightly coupled with the underlying scene representation. Earlier methods relied on pointcloud schemes such as octree partitioning and wavelet transforms, forming the basis of standards like MPEG-PCC~\cite{gpcc}. While efficient for static geometry, these approaches struggle with fidelity and compression ratio for dynamic content. 
Neural representations prompted learning-based compression for dynamic NeRF-style scenes~\cite{jointrf,hpc,VRVVC,MaskedWavelet}, often encoding residuals or optimizing compact, entropy-aware feature volumes.  However, their reliance on dense volumetric grids or sequential decoding pipelines often leads to limited rendering speed and high memory overhead, restricting their applicability in real-time scenarios. 
For 3D Gaussian Splatting (3DGS), most compression efforts have focused on static scenes~\cite{Lightgaussian,hac2024,contextgs,3DGSzip}, while dynamic extensions~\cite{4DGC,v3,dualgs} remain relatively underdeveloped. Existing dynamic 3DGS solutions often suffer from structural redundancy and rigid bitrate settings, offering limited control over resource allocation during transmission. 
Our approach addresses these limitations by reducing static background redundancy through motion decoupling and incorporating a lightweight, differentiable entropy model for end-to-end rate-aware optimization. This combination allows us to achieve improved RD performance while maintaining rendering efficiency and adaptability across diverse dynamic scenes.

\section{Motion Decoupled Representation}
\begin{figure*}[htb]
    \centering
    \includegraphics[width=1\linewidth]{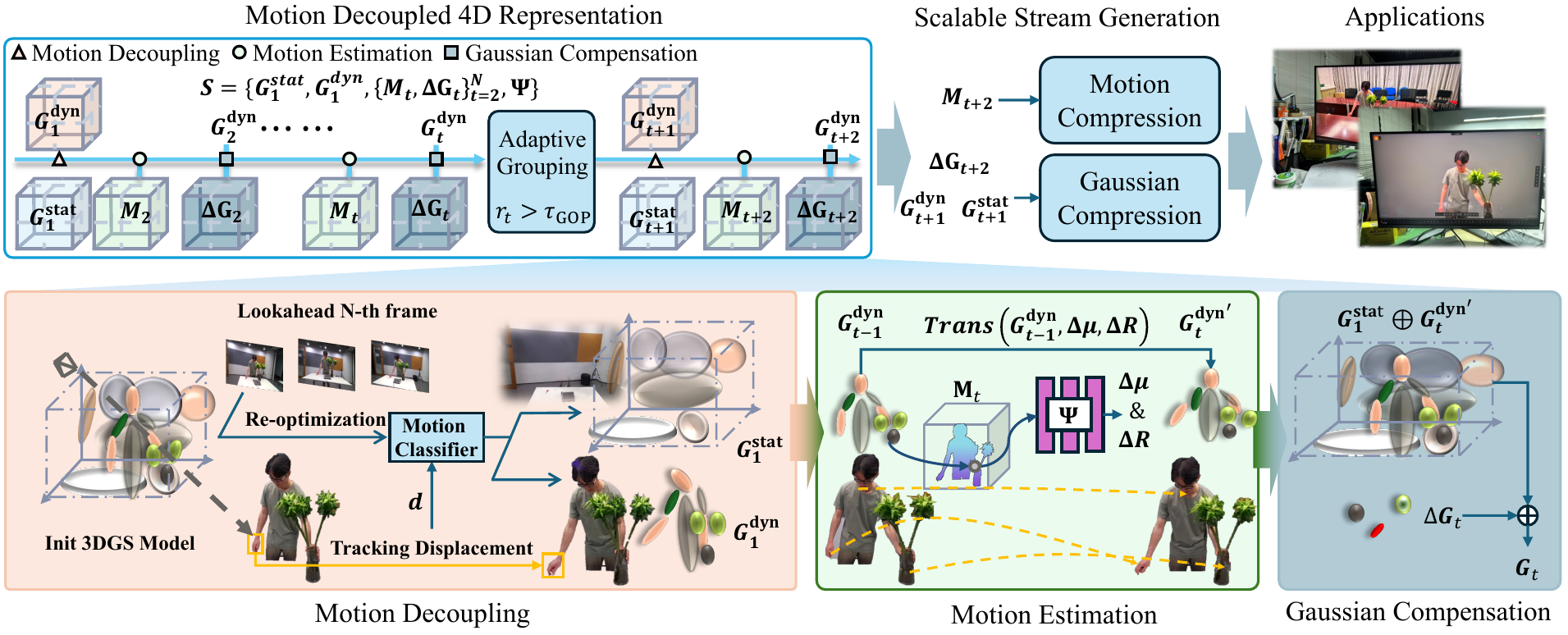}
    \caption{
    Overview of 4D-MoDe pipeline.
    Keyframe Gaussians are initially separated into static ($\mathbf{G}_1^{\text{stat}}$) and dynamic ($\mathbf{G}_1^{\text{dyn}}$) sets using a lookahead-based motion decoupling strategy. For subsequent inter-frames, dynamic Gaussians are transformed via a multi-resolution motion grid ($\mathbf{M}_t$) and a shared MLP ($\Psi$), and then refined with sparse compensated Gaussians ($\Delta \mathbf{G}_t$) to handle new details and form the complete frame representation ($\mathbf{G}_t$). 
    An adaptive grouping mechanism  manages keyframe insertion based on scene dynamics. Finally, all components, including motion grids and Gaussian sets, are independently compressed for scalable stream generation, supporting various applications.
    }
    \label{fig:pipeline}
\end{figure*}

To address the inherent limitation of traditional 3DGS in effectively distinguishing between static backgrounds and dynamic foregrounds, which often leads to modeling inefficiency and data redundancy, we propose a novel motion-decoupled representation for volumetric video (Fig.~\ref{fig:pipeline}). Our approach first explicitly separates dynamic foreground motion from static background elements through a lookahead-based motion decoupling strategy (Sec. \ref{sec:MotionDecoupling}). We then model the dynamic components using a multi-resolution motion estimation grid (Sec. \ref{sec:DynamicModeling}), complemented by Gaussian compensation to enhance visual quality (Sec. \ref{sec:MotionCompensation}). Furthermore, we introduce an adaptive GOP scheme that selectively inserts background keyframes only when significant scene transitions occur, thereby optimizing storage efficiency while maintaining reconstruction fidelity (Sec. \ref{sec:adaptivegrouping}).

\subsection{Overview}
3D Gaussian Splatting (3DGS) models a scene as a set of $N$ explicit Gaussian primitives $\mathbf{G} = \{\boldsymbol{\mathcal{G}}_i\}_{i=1}^N$. Each primitive $\boldsymbol{\mathcal{G}}_i$ is defined by a 3D center $\boldsymbol{\mu}_i$, a rotation $\mathbf{R}_i \in SO(3)$, principal axis scales $\mathbf{s}_i \in \mathbb{R}^3$, spherical harmonic (SH) coefficients $\mathbf{f}_i$ for appearance, and an opacity $\alpha_i$. The 3D covariance matrix $\boldsymbol{\Sigma}_i$ is constructed from rotation and scale as:
\begin{equation}
\boldsymbol{\Sigma}_i = \mathbf{R}_i \mathbf{S}_i \mathbf{S}_i^\top \mathbf{R}_i^\top,
\end{equation}
where $\mathbf{S}_i = \mathrm{diag}(\mathbf{s}_i)$ is a diagonal matrix of the scales. During rendering, these 3D Gaussians are projected onto the 2D image plane. This involves transforming from world to camera coordinate with extrinsic matrix $\mathbf{W}$ and computing the projected 2D covariance by the Jacobian $\mathbf{J}$ of the affine approximation of the projective transformation:
\begin{equation}
\boldsymbol{\Sigma}_i' = \mathbf{J} \mathbf{W} \boldsymbol{\Sigma}_i \mathbf{W}^\top \mathbf{J}^\top.
\end{equation}
The projected 2D Gaussians are then sorted by depth and rendered image $\hat{I}$ using alpha blending in \textit{z}-buffer. This process enables efficient, differentiable, and photorealistic image synthesis from the sparse Gaussian primitives.

To enable efficient modeling and structured editability, we represent the 4D scene as a motion-decoupled Gaussian sequence. At the keyframe ($t=1$), we initialize the full scene with a static background $\mathbf{G}_1^{\text{stat}}$ and dynamic foreground $\mathbf{G}_1^{\text{dyn}}$. For each subsequent frame $t > 1$, the dynamic Gaussians $\mathbf{G}_t^{\text{dyn}}$ are derived by applying rigid transformations estimated via a multi-resolution motion field $\mathbf{M}_t$, predicted by a shared MLP $\Psi$. To handle emerging or highly dynamic content, a sparse set of compensated Gaussians $\Delta \mathbf{G}_t$ is added.
Our decoupled 4D Gaussian representation is :
\begin{align}
\mathcal{S} = \left\{ \mathbf{G}_1^{\text{stat}},\; \mathbf{G}_1^{\text{dyn}},\; \{ \mathbf{M}_t,\; \Delta \mathbf{G}_t \}_{t=2}^{N},\; \Psi \right\},    
\end{align}
where the static Gaussians remain unchanged, and the dynamic content evolves over time via motion-guided transformations and adaptive compensation. This factorized representation supports scalable streaming and enables downstream editing such as background replacement and foreground-only rendering.

\subsection{Motion Decoupling}\label{sec:MotionDecoupling}
A critical aspect of our representation is the initial, robust separation of the keyframe into its static ($\mathbf{G}_1^{\text{stat}}$) and dynamic ($\mathbf{G}_1^{\text{dyn}}$) components. To achieve this separation without relying on external priors or segmentation models, we propose a fully learnable, multi-step strategy. This strategy derives motion cues by observing attribute changes over a short future interval, effectively performing a 'lookahead' to inform the decoupling process. It leverages the inherent fitting capabilities of 3DGS \cite{3dgs} to infer motion, as detailed below.

The process starts with a 3DGS model ($\mathbf{G}_1$) trained on the initial frame ($t=1$). A brief optimization adjustment is then performed on a subsequent frame ($t_1$, typically a short interval after the keyframe, as detailed in Sec.~\ref{sec:opt_decoupling}), allowing only Gaussian position ($\boldsymbol{\mu}$) and scale ($\mathbf{s}$) to change while keeping the Gaussian count fixed. These attribute adjustments form the basis for initial classification. Specifically, the motion offset of each Gaussian $i$ is examined via its image-space projected displacement $\mathbf{d}_i$, normalized by its depth $z_i$, denoted as $\delta_i = \frac{\|\mathbf{d}_i\|_2}{z_i}$. Concurrently, its scale change $\| \Delta \mathbf{s}_i \|_2$ is also considered. The magnitudes of $\delta_i$ and $\| \Delta \mathbf{s}_i \|_2$, when compared against preset thresholds $\tau_{\delta}$ and $\tau_{s}$ respectively, jointly serve as the initial basis for distinguishing dynamic and static points. Gaussians exceeding either threshold are initially classified as dynamic ($\mathbf{G}^{\text{dyn-pre}}$).

To enhance classification accuracy and filter out artifact points that might exhibit large changes due to poor fitting rather than true motion, we further introduce a refinement mechanism based on L1 rendering error. This mechanism operates by evaluating each Gaussian $\boldsymbol{\mathcal{G}}_i$'s weighted contribution $\mathcal{E}_{L1_i}$ to the rendering L1 error ($q_k = \| I_k - \hat{I}_k \|_1$ at pixel $k$), calculated as:
\begin{equation}
\mathcal{E}_{L1_i} = \frac{ \sum_{k} (C_{i,k} \times q_k) }{ \sum_{k} C_{i,k} },
\label{eq:e_l1}
\end{equation}
where $C_{i,k}$ represents the alpha blending weight of Gaussian $i$ at pixel $k$. If an initially classified dynamic point exhibits an $\mathcal{E}_{L1_i}$ exceeding a preset error threshold $\tau_{E}$, it is reclassified as static, under the assumption that high-change, high-error points are likely fitting artifacts.

Finally, to ensure the spatial regularity and physical plausibility of the separation results, we apply spatial clustering-based smoothing. This step is based on the premise that adjacent Gaussians in 3D space should share the same motion properties (either static or dynamic), aiming to enforce local consistency. It smooths the dynamic-static boundaries and eliminates isolated clusters, thus preventing rendering artifacts such as holes or floaters. Specifically, it operates through a K-Nearest Neighbors (KNN) based majority voting mechanism, adjusting each Gaussian's class according to the dominant class within its 3D spatial neighborhood, thus effectively integrating the scene's spatial structure information into the classification. Through this multi-step strategy of 'Initial Classification, Error Refinement, Clustering Smoothing', we generate the high-quality and structurally sound static ($\mathbf{G}_1^{\text{stat}}$) and dynamic ($\mathbf{G}_1^{\text{dyn}}$) Gaussian sets required for our keyframe representation. This provides a solid foundation for the subsequent motion estimation and compensation processes.

\subsection{Motion Estimation} \label{sec:DynamicModeling}
To model the temporal motion of dynamic Gaussian primitives, we propose a motion estimation module based on a hierarchical motion grid and a lightweight shared MLP. Our goal is to predict frame-wise translations and rotations for each Gaussian in $\mathbf{G}_{t-1}^{\text{dyn}}$ to achieve temporally coherent updates.

We represent the motion field for frame $t$ using a set of multi-resolution volumetric grids $\mathbf{M}_t = \{\mathbf{M}_t^l\}_{l=1}^L$, where each grid $\mathbf{M}_t^l$ captures motion features at spatial level $l$. For each Gaussian $\boldsymbol{\mathcal{G}} \in \mathbf{G}_{t-1}^{\text{dyn}}$, we extract its center $\boldsymbol{\mu}_{t-1}$ and apply a positional encoding $\gamma(\boldsymbol{\mu}_{t-1})$ to obtain multi-scale embeddings $\gamma^l_{t-1}$. At each resolution level $l$, we perform trilinear interpolation on the corresponding grid $\mathbf{M}_t^l$ using the encoded position to extract motion features. The interpolated features from all levels are then concatenated (denoted by $\operatorname{concat}(\cdot)$) to form a unified motion descriptor:
\begin{equation}
\mathbf{f}_{\text{motion}}(\boldsymbol{\mu}_{t-1}) = \operatorname{concat}_{l=1}^L \left( \text{interp}\left(\gamma^l_{t-1}, \mathbf{M}_t^l\right) \right),
\end{equation}
where $\text{interp}(\cdot)$ denotes the trilinear interpolation operation.

This motion descriptor is then fed into a lightweight shared MLP, $\Psi$, to predict the per-Gaussian translation and rotation increments:
\begin{equation}
 \Delta \boldsymbol{\mu}_t= \Psi(\mathbf{f}_{\text{motion}}), \quad \Delta \mathbf{R}_t= \Psi(\mathbf{f}_{\text{motion}}) .
\end{equation}
The MLP $\Psi$ is designed to output both the translation vector $\Delta \boldsymbol{\mu}_t$ and a quaternion representation for the rotation increment, which is subsequently converted to the matrix $\Delta \mathbf{R}_t$. The updated Gaussians for frame $t$ are then obtained by applying these increments to the previous frame’s Gaussians:
\begin{align}
& \mathbf{G}_t^{\text{dyn}'} = \textbf{Trans}(\mathbf{G}_{t-1}^{\text{dyn}},\Delta \boldsymbol{\mu}_t,\Delta \mathbf{R}_t) \\
&= \left\{ \boldsymbol{\mathcal{G}}(\boldsymbol{\mu}_{t-1} + \Delta \boldsymbol{\mu}_t;\; \Delta \mathbf{R}_t  \mathbf{R}_{t-1};\; C) \;\big|\; \boldsymbol{\mathcal{G}} \in \mathbf{G}_{t-1}^{\text{dyn}} \right\},
\end{align}
where $C$ represents fixed attributes such as scale $\mathbf{s}$, appearance coefficients $\mathbf{f}$, and opacity $\alpha$ from the previous frame. This hierarchical motion estimation approach enables accurate and scalable modeling of complex dynamic content by capturing motion patterns across multiple spatial resolutions.

\subsection{Gaussian Compensation} \label{sec:MotionCompensation}
Although the motion estimation module effectively models frame-to-frame transformations for previously observed dynamic objects, it can struggle with fine-grained dynamics, sudden object emergence, or abrupt motion changes. These situations can lead to visual discrepancies that require correction. To mitigate these limitations, we introduce a Gaussian compensation mechanism. This mechanism augments the scene representation by adding a sparse set of compensated Gaussians, $\Delta \mathbf{G}_t$, which are optimized to enhance visual quality and capture details missed by the predictive motion model, thereby improving the overall expressiveness of the dynamic scene representation. We identify regions for compensation based on two primary criteria:

First, to handle newly emerging content or areas with significant appearance changes (often indicated by high image-space gradients), we replicate relevant Gaussians. Specifically, pixels with image-space gradients $\|\nabla I_t(\pi(\boldsymbol{\mu}))\|$ (where $\pi(\cdot)$ is the projection) exceeding a threshold $\tau_g$ are identified. Gaussians from the transformed dynamic set $\mathbf{G}_t^{\text{dyn}'}$ whose centers project to these high-gradient regions are replicated to form an initial compensated set $\Delta \mathbf{G}_t^g$:
\begin{align}
\Delta \mathbf{G}_t^g = \left\{ \boldsymbol{\mathcal{G}} \in \mathbf{G}_t^{\text{dyn}'} \;\middle|\; \|\nabla I_t(\pi(\boldsymbol{\mu}))\| > \tau_g \right\}.
\end{align}
Second, for dynamic Gaussians undergoing large transformations, we perform Gaussian splitting to better capture localized motion. This is triggered if the translation magnitude $\|\Delta \boldsymbol{\mu}_t\|$ exceeds $\tau_{\mu}$, or if the rotation deviation (derived from the quaternion representation of $\Delta \mathbf{R}_t$) exceeds $\tau_R$. In such cases, the original Gaussian is replaced by two child Gaussians sampled from a local region $\mathcal{N}(\boldsymbol{\mu}, 2\boldsymbol{\Sigma})$, each initialized with significantly downscaled scales:
\begin{align}
    \mathbf{s}' = \frac{\mathbf{s}}{100}.
\end{align}
Let $\Delta \mathbf{G}_t^m$ denote this set of motion-based compensated (split) Gaussians. The full set of compensated Gaussians for the current frame is then $\Delta \mathbf{G}_t = \Delta \mathbf{G}_t^g \cup \Delta \mathbf{G}_t^m$.

These newly generated Gaussians are incorporated into the dynamic part of the scene representation for the current frame:
\begin{equation}
\mathbf{G}_t^{\text{dyn}} = \mathbf{G}_t^{\text{dyn}'} \cup \Delta \mathbf{G}_t.
\end{equation}
The complete representation for rendering frame $t$ then consists of the static background $\mathbf{G}_1^{\text{stat}}$ and this updated dynamic set $\mathbf{G}_t^{\text{dyn}}$. During a dedicated training stage (Sec.~\ref{sec:opt_compensation}), the appearance and opacity attributes of $\Delta \mathbf{G}_t$ are optimized. This compensation strategy enhances spatial detail, mitigates motion blur, and ensures temporal consistency, serving as a crucial complement to the motion estimation process for achieving high-quality reconstruction of dynamic scenes.

\subsection{Adaptive Grouping} \label{sec:adaptivegrouping}
Traditional volumetric video modeling methods often utilize a fixed GOP size for compression and encoding. However, fixed GOP structures are ill-suited for dynamic scenes, especially those with complex or rapidly changing backgrounds. They often fail to promptly reflect scene changes in the static background model, leading to degraded reconstruction quality and increased redundancy. For instance, excessively long GOPs struggle to adapt to emergent or disappearing content in static regions, degrading quality over time. Conversely, overly short GOPs lead to frequent keyframe insertion, significantly increasing the overall model size and transmission overhead. To address these issues, we propose an adaptive GOP selection strategy based on the frame-wise dynamics of the scene. Specifically, at each frame $t$, we compute the ratio of newly introduced dynamic Gaussians  to the total dynamic Gaussians from the previous frame as:
\begin{align}
r_t = \frac{|\Delta \mathbf{G}_t|}{|\mathbf{G}_{t-1}^{\text{dyn}}|},
\end{align}
where $|\cdot|$ denotes the number of Gaussians. When this ratio $r_t$ exceeds a predefined threshold $\tau_{\text{GOP}}$, a new GOP is initiated. This criterion ensures that the background model is refreshed only when substantial changes occur, balancing stability and adaptability.

When a new GOP is initiated, a new keyframe is established to accurately capture the current scene state. To ensure temporal continuity and leverage prior information, the 3DGS model for this new keyframe is first initialized using the complete set of Gaussians (both static and dynamic) from the final frame of the preceding GOP. This initialized model then undergoes an optimization process against the multi-view images of the new keyframe. Subsequently, our full lookahead-based motion decoupling strategy is applied to this optimized keyframe model. This yields a new, distinct set of static background Gaussians ($\mathbf{G}_{\text{new\_kf}}^{\text{stat}}$) and dynamic foreground Gaussians ($\mathbf{G}_{\text{new\_kf}}^{\text{dyn}}$) tailored to the current keyframe. This approach allows the static background to be effectively refreshed and adapted to significant scene changes, while also re-establishing a clean separation of the dynamic foreground. The inter-frames within this new GOP then operate based on this newly defined keyframe structure.

\section{Progressive Multi-Stage Optimization} \label{sec:optimization}

\begin{figure*}[htb]
    \centering
    \includegraphics[width=\linewidth]{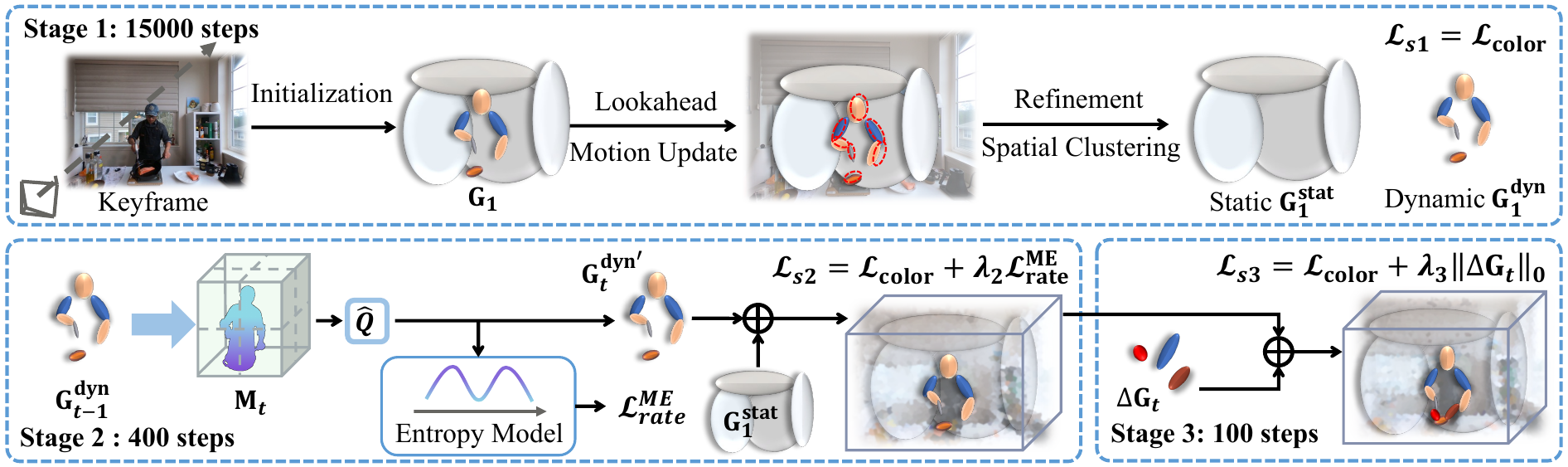}
    \caption{Overview of our progressive multi-stage optimization. First, we obtain initial Gaussians from keyframe and supervise the motion-decoupling process via photometric loss to divide Gaussians into $\mathbf{G}^{\text{stat}}_1$ and $\mathbf{G}^{\text{dyn}}_1$. For dynamic Gaussians at time $t$, we simulate quantization and perform bitrate estimation on motion grid to derive $\mathcal{L}_{\text{rate}}^{\text{ME}}$, which is combined with the photometric term to supervise the motion estimation process. Finally, we conduct Gaussian compensation and introduce supervision on the number of compensated Gaussians.}
    \label{fig:training}
\end{figure*}

To learn a 4D-MoDe representation that is both precise in depicting dynamic scenes and highly amenable to compression, we propose a progressive multi-stage optimization framework, illustrated in Fig.~\ref{fig:training}. This framework systematically constructs and refines our motion-decoupled 4D Gaussian representation through three carefully designed, sequential stages. The first stage (Sec.~\ref{sec:opt_decoupling}) focuses on establishing the scene's core motion-decoupled structure. The second stage (Sec.~\ref{sec:opt_motion}) introduces RD constraints, emphasizing the optimization of the motion field's representation efficiency. The final stage (Sec.~\ref{sec:opt_compensation}) concentrates on maximizing reconstruction fidelity via Gaussian compensation. To enable differentiable RD optimization for motion, we integrate a Simulated Quantization strategy and a lightweight Implicit Entropy Model, which are detailed in Sec.~\ref{sec:opt_motion}.

\subsection{Motion Decoupling Optimization} \label{sec:opt_decoupling}

This foundational stage implements the multi-step motion decoupling strategy, as detailed in Sec.~\ref{sec:MotionDecoupling}, to partition the initial Gaussians into static ($\mathbf{G}_1^{\text{stat}}$) and dynamic ($\mathbf{G}_1^{\text{dyn}}$) sets. The procedure involves two key optimization phases followed by classification and refinement steps, all designed to establish an accurate structural foundation for subsequent stages. Initially, we perform a standard 3D Gaussian Splatting training procedure utilizing only the multi-view images ($t=1$) of the first frame of input sequence. During this process, all Gaussian attributes are optimized to minimize the $\mathcal{L}_{\text{color}}$ loss. This loss function is a weighted sum of the L1 loss and the D-SSIM loss \cite{SSIM} between the rendered image and the ground truth frame $t=1$. The result is a high-fidelity 3DGS model, denoted as $\mathbf{G}_1$, representing the initial state of the scene.
\begin{align}
    \mathcal{L}_{\text{s1}} = \mathcal{L}_{\text{color}}= \lambda_1 \|I - \hat{I}  \|_1 + (1-\lambda_1) \mathcal{L}_{\text{D-SSIM}}
\end{align}

Subsequently, we conduct a brief fine-tuning phase using a later frame ($t_1$, typically chosen around the 30th frame). Starting with the parameters of $\mathbf{G}_1$, we freeze the Gaussian count, rotation, appearance, and opacity. Only the position ($\boldsymbol{\mu}$) and scale ($\mathbf{s}$) attributes are allowed to be optimized. This phase runs for a shorter duration, minimizing the $\mathcal{L}_{\text{color}}$ loss, but this time against the ground truth frame at $t_1$. This targeted optimization encourages Gaussians located in regions undergoing motion to adjust their $\boldsymbol{\mu}$ and $\mathbf{s}$, thereby generating the critical signals required for effective decoupling.

Upon completion of these two optimization phases, the classification and refinement pipeline is executed. For each Gaussian, we calculate its normalized projected displacement $\delta_i$ by tracking its position change between $\mathbf{G}_1$ and the fine-tuned model, and its scale change $\| \Delta \mathbf{s}_i \|_2$. An initial classification ($\mathbf{G}^{\text{dyn-pre}}$) is performed by applying the thresholds $\tau_{\delta}$ and $\tau_{s}$. Following this, we compute the per-Gaussian L1 error contribution $\mathcal{E}_{L1_i}$ (Eq.~\ref{eq:e_l1}) using the final state of the fine-tuned model against frame $t_1$. Gaussians initially marked as dynamic but exhibiting an error $\mathcal{E}_{L1_i}$ above the threshold $\tau_{E}$ are reclassified as static. Finally, a K-Nearest Neighbors (KNN) based spatial clustering and majority voting is applied in 3D space to enforce local consistency and smooth the boundaries.
This stage is to produce accurate and structurally sound $\mathbf{G}_1^{\text{stat}}$ and $\mathbf{G}_1^{\text{dyn}}$ sets, which are essential inputs for the subsequent RD focused optimization stages.

\subsection{Motion Estimation Optimization} \label{sec:opt_motion}
Given the dynamic Gaussians $\mathbf{G}_{t-1}^{\text{dyn}}$ from the previous frame, this stage focuses on optimizing $\mathbf{M}_t$ and $\Psi$ to accurately model foreground motion while maintaining compressibility. The objective is to achieve effective RD performance, ensuring both precise motion representation and low entropy for the motion features.

To integrate RD objectives into a differentiable optimization pipeline, we adopt two key techniques. First, we introduce simulated quantization by injecting uniform noise $u \sim U\left(-\frac{1}{2q}, \frac{1}{2q}\right)$ to mimic quantization effects with step size $q$, while preserving gradient flow. Second, we employ a lightweight, trainable implicit entropy model to estimate the probability mass function (PMF) of quantized motion features $\hat{y}$ via a learned cumulative distribution function (CDF):
\begin{align}
P_{\text{PMF}}(\hat{y}) = P_{\text{CDF}}\left(\hat{y} + \tfrac{1}{2}\right) - P_{\text{CDF}}\left(\hat{y} - \tfrac{1}{2}\right). \label{eq:pmf_from_cdf}
\end{align}
This enables on-the-fly bitrate estimation and entropy supervision compatible with backpropagation. During training, the appearance, opacity, and scale parameters of all Gaussians are fixed. We jointly optimize the motion grid $\mathbf{M}_t$, the MLP $\Psi$, and the entropy model. The motion grid is first processed with simulated quantization, producing $\hat{\mathbf{M}}_t$ for bitrate estimation. The overall loss function $\mathcal{L}_{\text{s2}}$ is defined as:
\begin{equation}
\mathcal{L}_{\text{s2}} = \mathcal{L}_{\text{color}} + \lambda_2 \mathcal{L}_{\text{rate}}^{\text{ME}},
\label{eq:ls2_motion_opt}
\end{equation}
where $\mathcal{L}_{\text{rate}}^{\text{ME}}$ estimates the bitrate of the quantized motion features $\hat{\mathbf{M}}_t$:
\begin{align}
\mathcal{L}_{\text{rate}}^{\text{ME}} = -\frac{1}{N} \sum_{\hat{y} \in \hat{\mathbf{M}}_t} \log_2\left( P_{\text{PMF}}(\hat{y}) \right). \label{eq:rate_loss_me_def_actual}
\end{align}
The output of this stage is an optimized motion grid $\mathbf{M}_t$ and its corresponding entropy model, providing a temporally coherent and compressible motion representation.

\subsection{Gaussian Compensation Optimization} \label{sec:opt_compensation}
The final optimization stage is dedicated to refining the compensated Gaussians, $\Delta \mathbf{G}_t$. Its purpose is to enhance the scene representation by capturing fine details, modeling newly emerged content, or addressing complex deformations that were missed by the motion estimation stage, thereby maximizing the final reconstruction quality. We start with a base representation for the current frame, formed by combining the static background $\mathbf{G}_1^{\text{stat}}$ with the motion-predicted dynamic Gaussians $\mathbf{G}_t^{\text{dyn}'}$.
To this base, a sparse set of compensated Gaussians, $\Delta \mathbf{G}_t$, is introduced; their generation strategy is detailed in Sec.~\ref{sec:MotionCompensation}. All attributes of these newly added $\Delta \mathbf{G}_t$ are subsequently optimized under the guidance of the loss function $\mathcal{L}_{\text{s3}}$:
\begin{equation}
    \mathcal{L}_{\text{s3}} = \mathcal{L}_{\text{color}} + \lambda_3 \|  \Delta \mathbf{G}_t \|_0,
    \label{eq:ls3_comp_opt}
\end{equation}
where $\mathcal{L}_{\text{color}}$ measures the reconstruction error of the complete scene (including $\mathbf{G}_1^{\text{stat}}$, $\mathbf{G}_t^{\text{dyn}'}$, and the optimized $\Delta \mathbf{G}_t$) against the ground truth, and the term $\lambda_3 \| \Delta \mathbf{G}_t \|_0$ regularizes the number of compensated Gaussians, promoting sparsity and ensuring the compactness of the representation. Minimizing $\mathcal{L}_{\text{s3}}$ thus balances the goals of accurately correcting rendering errors and enhancing details with the need to maintain a concise set of compensated Gaussians, ultimately yielding an optimized and efficient $\Delta \mathbf{G}_t$.

Through these three progressive optimization stages, our 4D-MoDe framework learns a hierarchical and efficient dynamic scene representation. Motion information is compressed via training-time RD optimization, while scene structure and compensation details are refined for high-quality reconstruction and subsequently encoded, achieving an excellent balance between performance and flexibility.

\section{Scalable Stream Generation}
\label{sec:stream_generation}

\subsection{Motion Field Compression}
\label{sec:motion_compression}

Upon completing the training process, the multi-resolution motion field $\mathbf{M}_t$ for each inter-frame requires efficient compression to generate the final bitstream. We employ a strategy combining quantization and range coding for entropy encoding. While our training utilizes an implicit entropy model (as part of $\Psi$), transmitting its full parameters can be suboptimal. Instead, we analyze the data distribution $\omega_t^\mathbf{M}$ of the quantized motion field prior to encoding and transmit this compact distribution representation alongside the encoded data, effectively reducing overhead.

The encoding process for the motion field $\mathbf{M}_t$ involves quantization $\mathbf{Q}$ followed by range encoding $\mathbf{E}$. The quantization step is defined as:
\begin{equation}
    \mathbf{Q}(\mathbf{M}_t) = \left\lfloor q \cdot \mathbf{M}_t + 0.5 \right\rfloor,
\end{equation}
where $q$ is the quantization step size. The range encoder $\mathbf{E}$ then produces the bitstream $B_t^\mathbf{M}$:
\begin{equation}
    B_t^\mathbf{M} = \mathbf{E} \left( \mathbf{Q}(\mathbf{M}_t) - \mathbf{Q}(\min(\mathbf{M}_t)); \omega_t^\mathbf{M} \right).
    \label{eq:motion_encode}
\end{equation}
Here, we subtract the quantized minimum value to ensure non-negative inputs for the range encoder.

For efficient storage, such as in the int8 format, the quantized values are mapped to non-negative integers. The quantization parameter $q$ serves to control the balance between precision and the resulting range of values. The decoding process involves range decoding $\mathbf{D}$ followed by dequantization:
\begin{equation}
    \hat{\mathbf{M}}_t = \frac{\mathbf{D}\left(B_t^\mathbf{M}; \omega_t^\mathbf{M} \right) + \mathbf{Q}(\min(\mathbf{M}_t))}{q}.
    \label{eq:motion_decode}
\end{equation}
Thus, for each inter-frame's motion component, we only need to transmit the motion bitstream $B_t^\mathbf{M}$ and its corresponding distribution representation $\omega_t^\mathbf{M}$.

\subsection{Gaussian Compression} \label{sec:gaussian_compression}
All Gaussian primitives required for rendering—the initial static background ($\mathbf{G}_1^{\text{stat}}$), initial dynamic foreground ($\mathbf{G}_1^{\text{dyn}}$), and all per-frame compensated Gaussians ($\Delta \mathbf{G}_t$)—are compressed using a unified framework leveraging Google's Draco \cite{draco}. In our adaptation, the various Gaussian attributes are separated by channel and integrated as geometric meta-attributes within Draco's KD-tree spatial partitioning structure. These attribute channels subsequently undergo quantization and are entropy-encoded using Draco's range coding back-end. This comprehensive pipeline achieves high compression ratios while preserving visual fidelity, producing efficient streams suitable for storage and transmission in volumetric video streaming systems; notably, this Draco-based encoding for Gaussians typically completes in under 10ms, with decoding achieved in under 5ms.

\subsection{Applications}
The proposed motion-decoupled 4D Gaussian representation not only achieves high compression efficiency and rendering quality, but also significantly enhances the editability and flexibility of volumetric video systems. By explicitly separating static backgrounds from dynamic foregrounds, our method enables flexible editing operations such as background replacement. Without the need for re-capturing content, users can modify or substitute the background directly during the rendering stage, facilitating seamless transitions from real-world environments to virtual spaces. This capability is particularly useful for applications such as virtual conferencing, immersive broadcasting, and visual effects production.

\begin{figure}[htb]
    \centering
    \includegraphics[width=0.95\linewidth]{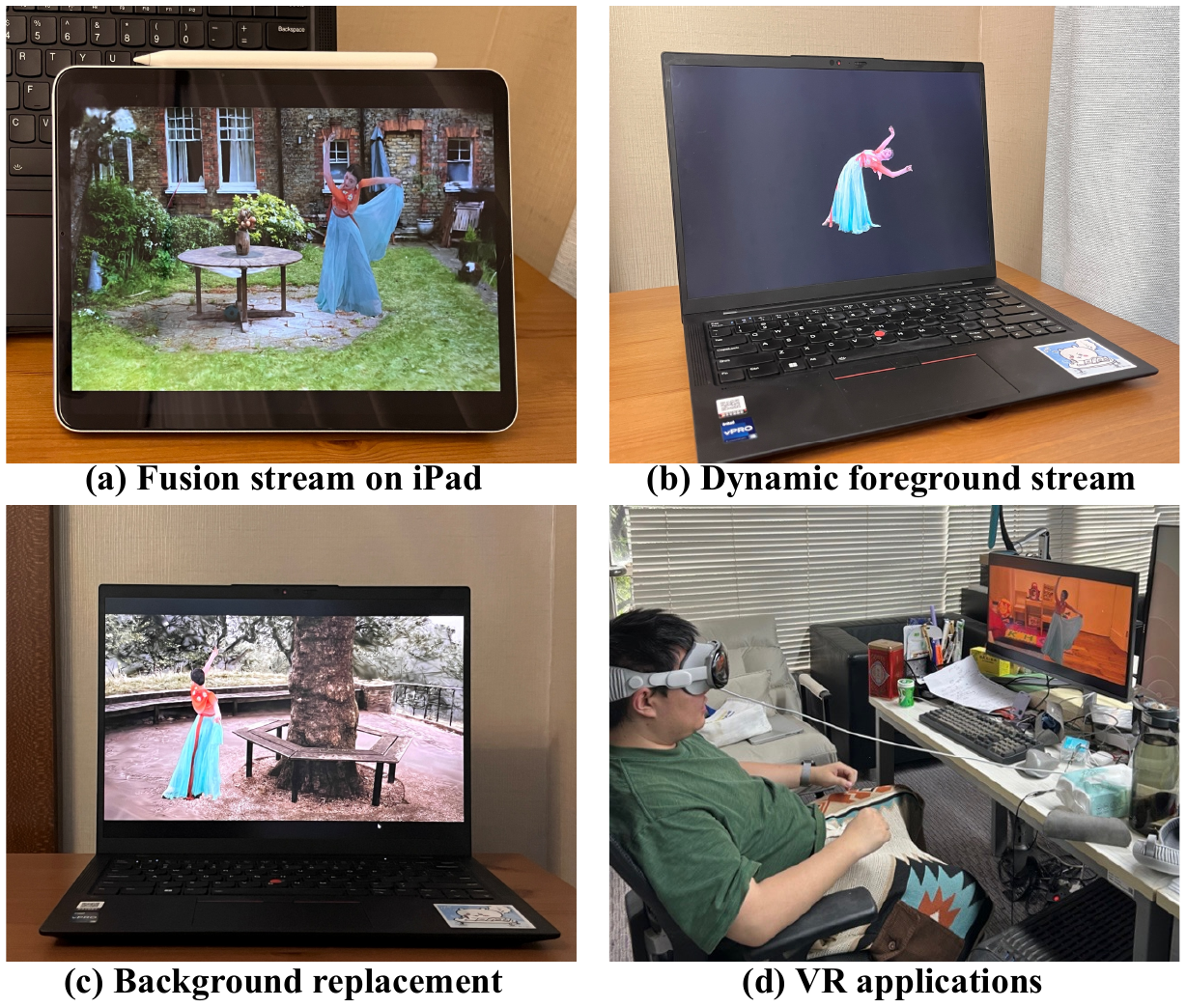}
    \caption{Demonstration of 4D-MoDe's diverse application capabilities, enabled by its motion-decoupled framework. Use cases include: (a) volumetric fusion stream on mobile devices (iPad), (b) dynamic foreground streaming, (c) flexible background replacement for content editing, and (d) interactive virtual reality (VR) experiences.}
    \label{fig:applications}
\end{figure}

Moreover, the decoupled structure supports foreground-only streaming, where only the dynamic foreground Gaussians are encoded and transmitted over the network, while the static background can be preloaded or reused at the receiver side. This design reduces transmission bandwidth and supports low-latency rendering, which is crucial for real-time communication and interactive systems. As illustrated in Fig.~\ref{fig:applications}, we showcase representative results including background replacement and independent foreground rendering, demonstrating the practical value of our method in building scalable and editable volumetric video applications.

\section{Experiments}

\subsection{Configurations}
\textbf{Datasets.} We validate the effectiveness of 4D-MoDe on three datasets: the N3DV dataset \cite{n3dv}, the MeetRoom dataset \cite{StreamRF}, and the Google Immersive dataset \cite{googleimmersive}. The N3DV dataset comprises 21 camera views at a resolution of $2704 \times 2028$, the MeetRoom dataset includes 14 views at $1280 \times 720$, and the Google Immersive dataset provides 46 views at $2560 \times 1920$. For each dataset, one camera view is held out for testing, while the remaining views are used for training.

\textbf{Implementation}.All experiments are conducted on Intel E5-2699 v4 CPU and a RTX 4090 GPU. For the multi-resolution motion grid $\mathbf{M}_t$, we use $L=3$ levels with feature channels of 4, 4, and 2, and corresponding grid dimensions of $32^3, 64^3$, and $128^3$, respectively.
Motion decoupling employs initial classification thresholds $\tau_{\delta}=4.5$ and $\tau_s=0.1$, an error refinement threshold $\tau_E=0.02$, and $K=5$ for K-Nearest Neighbors (KNN) smoothing.
Gaussian compensation utilizes $\tau_g = 0.0001,\tau_{\mu} = 0.08,\tau_R = \pi/4$ thresholds. The adaptive GOP threshold $\tau_{\text{GOP}}$ is set to $0.15$.
All Gaussian optimization processes use the Adam optimizer with a learning rate of $5 \times 10^{-4}$. For rate-distortion (RD) aware motion estimation, $\lambda_2$ is varied across $\{5e-5, 1e-4, 2e-4, 3e-4, 5e-4\}$ to achieve different bitrates. The loss weights for Stage 1 and Stage 3 are $\lambda_1=0.2$ and $\lambda_3=0.00005$, respectively.

\textbf{Metrics.} To assess the compression performance of our approach, we adopt Peak Signal-to-Noise Ratio (PSNR) and Structural Similarity Index Measure (SSIM) \cite{SSIM} as objective quality indicators, while bitrate is reported in kilobytes per frame (KB/frame). For a comprehensive evaluation of RD performance, we employ Bjontegaard Delta Bit-Rate (BDBR) and Bjontegaard Delta PSNR (BD-PSNR) metrics \cite{BDPSNR}. In addition, rendering efficiency is evaluated based on the number of frames rendered per second (FPS).

\subsection{Comparison}
\textbf{Quantitative comparisons.} To evaluate the performance of our proposed method, we conduct comprehensive comparisons against a range of state-of-the-art baselines including K-Planes~\cite{kplanes}, MixVoxels~\cite{MixVoxels}, HyperReel~\cite{attal2023hyperreel}, StreamRF~\cite{StreamRF}, ReRF~\cite{rerf}, TeTriRF~\cite{tetrirf}, HiCoM~\cite{hicom2024}, 4DGS~\cite{4dgs}, 3DGStream~\cite{3dgstream}, and 4DGC~\cite{4DGC}. As shown in Tab.~\ref{tab:n3dv_comparison}, our method achieves near-identical reconstruction quality to 4DGC (31.56\,dB/0.942 vs. 31.58\,dB/0.943) with only \textbf{18.3\%} of its storage (\textbf{93.5\,KB} vs. \textbf{511\,KB}), translating to a \textbf{5.46$\times$} compression ratio. Compared to other high-performance baselines like 3DGStream~\cite{3dgstream} (8294\,KB) and HiCoM~\cite{hicom2024} (716\,KB), our approach maintains superior visual quality while reducing the bitrate by \textbf{98.8\%} and \textbf{86.9\%}, respectively. Unlike streaming-oriented methods (e.g., ReRF~\cite{rerf}, TeTriRF~\cite{tetrirf}) that trade off quality for bandwidth adaptability, our solution uniquely combines: (1) state-of-the-art RD efficiency, (2) variable bitrate support, and (3) streaming-compatible capability. To further validate the generalizability of our method, we evaluate it on the MeetRoom~\cite{StreamRF} and Google Immersive~\cite{googleimmersive} datasets. Tab.~\ref{tab:streamable_comparison} shows that 4D-MoDe consistently ranks among the top in reconstruction quality, while offering the most compact model size and maintaining real-time rendering at over 140 FPS. This demonstrates its suitability for bandwidth-constrained volumetric video applications.

\begin{table}[tb]
\centering
\caption{Quantitative comparison on the N3DV dataset. The PSNR, SSIM, size, and rendering speed are averaged over the whole 300 frames for each scene. S/V indicates whether the method supports streaming and variable bitrate}
\label{tab:n3dv_comparison}
\scalebox{1}{
\begin{tabular}{c|cccc|c}
\hline
Method & \begin{tabular}[c]{@{}c@{}}PSNR$\uparrow$\\(dB)\end{tabular} & SSIM$\uparrow$  & \begin{tabular}[c]{@{}c@{}}Render$\uparrow$\\(FPS)\end{tabular} & \begin{tabular}[c]{@{}c@{}}Size$\downarrow$\\(KB)\end{tabular} & \begin{tabular}[c]{@{}c@{}}S/V\end{tabular} \\ \hline
K-Planes\cite{kplanes}  & 29.91 & 0.920 & 0.15   & 1065  & \ding{56} / \ding{56} \\
MixVoxels\cite{MixVoxels}  & 30.80 & 0.931 & 16.7   & 1710  & \ding{56} / \ding{56} \\
HyperReel\cite{attal2023hyperreel} & 31.10 & 0.938 & 2.5    & 1228  & \ding{56} / \ding{56} \\
StreamRF\cite{StreamRF} & 28.85 & 0.912 & 10.1   & 17070 & \ding{51} / \ding{56} \\
ReRF\cite{rerf}  & 29.71 & 0.918 & 168    & 512  & \ding{51} / \ding{51} \\
TeTriRF\cite{tetrirf}  & 30.65 & 0.931 & 2.7    & 778  & \ding{51} / \ding{51} \\
HiCoM\cite{hicom2024}  & 31.17 & 0.939 & \textbf{274}    & 716  & \ding{51} / \ding{51} \\
4DGS\cite{4dgs} & 30.19 & 0.930 & 114    & 20674 & \ding{56} / \ding{56} \\
3DGStream\cite{3dgstream}  & 31.23 & 0.940 & \underline{215}    & 8294  & \ding{51} / \ding{56} \\
4DGC\cite{4DGC} & \textbf{31.58} & \textbf{0.943} & 168    & \underline{511}  & \ding{51} / \ding{51} \\
Ours   & \underline{31.56} & \underline{0.942} & 172    & \textbf{93.5}  & \ding{51} / \ding{51} \\
\hline
\end{tabular}
}
\end{table}

\begin{table}[tb]
\setlength{\tabcolsep}{2pt} 
\renewcommand{\arraystretch}{1.05} 
\centering
\caption{Comparison on MeetRoom and Google Immersive datasets.S/V INDICATES WHETHER THE METHOD SUPPORTS STREAMING AND VARIABLE BITRATE}
\label{tab:streamable_comparison}
\scalebox{1}{
\begin{tabular}{c|c|cccc|c}
\hline
Dataset & Method & \begin{tabular}[c]{@{}c@{}}PSNR$\uparrow$\\ (dB)\end{tabular} & SSIM$\uparrow$ & \begin{tabular}[c]{@{}c@{}}Size$\downarrow$\\ (KB)\end{tabular} & \begin{tabular}[c]{@{}c@{}}Render$\uparrow$\\ (FPS)\end{tabular} & \begin{tabular}[c]{@{}c@{}}S/V\end{tabular} \\
\hline
\multirow{7}{*}{MeetRoom 
} 
& StreamRF\cite{StreamRF}   & 26.71 & 0.913 & 8428 & 10  & \ding{51}/\ding{56} \\
& ReRF \cite{rerf}          & 26.43 & 0.911 & 645 & 3   & \ding{51}/\ding{51} \\
& TeTriRF \cite{tetrirf}    & 27.37 & 0.917 & 624 & 4   & \ding{51}/\ding{51} \\
& 3DGStream\cite{3dgstream} & 28.03 & 0.921 & 8407 & \textbf{288} & \ding{51}/\ding{56} \\
& 4DGC \cite{4DGC}          & \textbf{28.08} & \textbf{0.922} & 430 & 213 & \ding{51}/\ding{51} \\
& Ours                      & \underline{28.01} & \underline{0.920} & \textbf{71.7} & 210 & \ding{51}/\ding{51} \\
\hline
\multirow{7}{*}{\shortstack{Google\\ Immersive 
}} 
& StreamRF\cite{StreamRF}   & 28.14 & 0.929 & 10486 & 8   & \ding{51}/\ding{56} \\
& ReRF \cite{rerf}          & 27.75 & 0.928 & 952  & 1   & \ding{51}/\ding{51} \\
& TeTriRF\cite{tetrirf}     & 28.53 & 0.931 & 850  & 2   & \ding{51}/\ding{51} \\
& 3DGStream\cite{3dgstream} & 29.66 & 0.935 & 10578 & \textbf{199} & \ding{51}/\ding{56} \\
& 4DGC\cite{4DGC}           & \textbf{29.71} & \textbf{0.935} & \underline{625}  & \underline{145} & \ding{51}/\ding{51} \\
& Ours                      & \textbf{29.71} & \textbf{0.935} & \textbf{92.2}  & 140 & \ding{51}/\ding{51} \\
\hline
\end{tabular}
}
\end{table}

\begin{figure}
    \centering
    \includegraphics[width=1\linewidth]{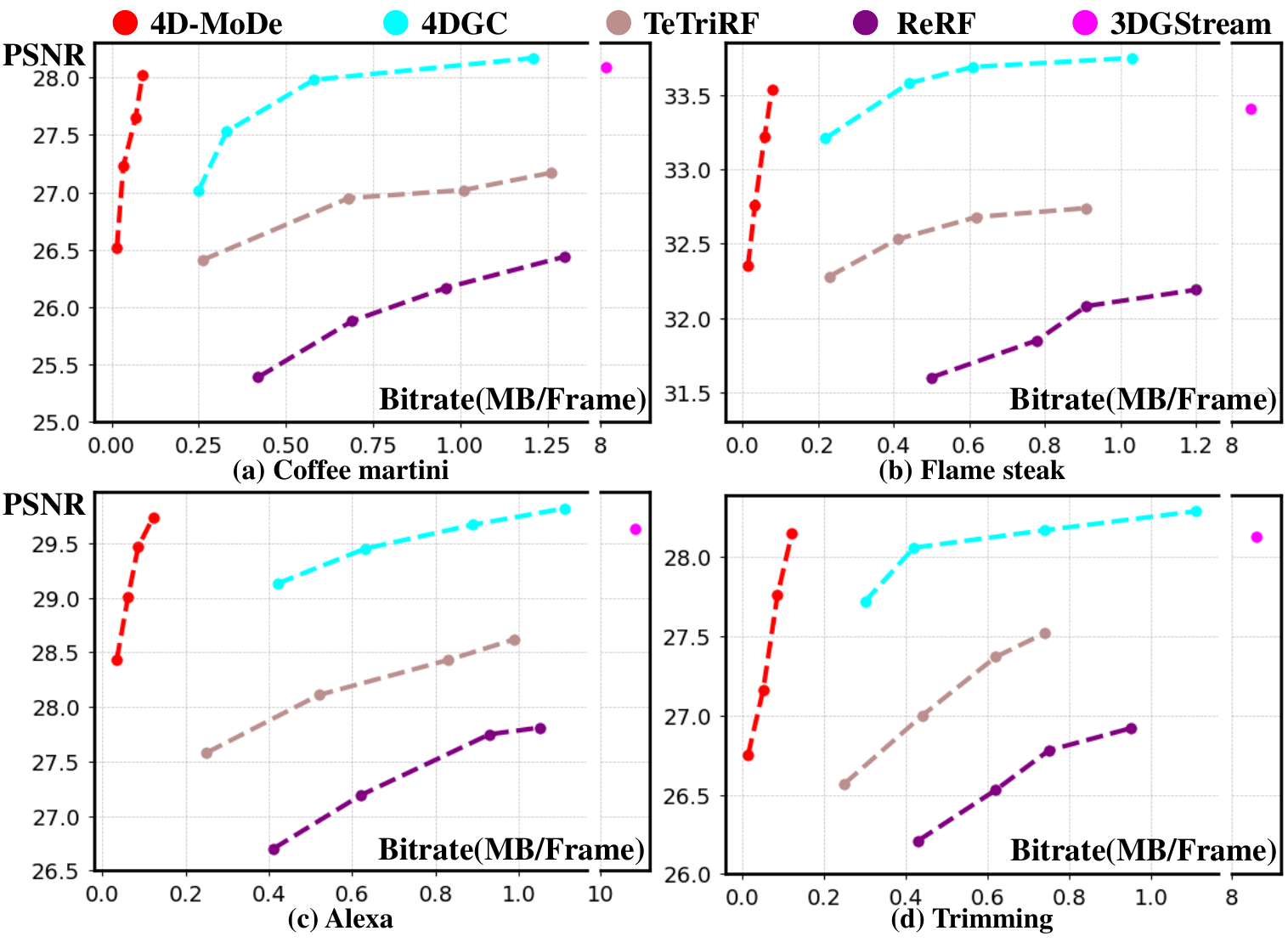}
    \caption{Rate-distortion curves across different datasets, showing the superior performance of our 4D-MoDe over ReRF \cite{rerf}, TeTriRF \cite{tetrirf}, 3DGStream \cite{3dgstream} and 4DGC\cite{4DGC}.}
    \label{fig:rd_curves}
\end{figure}

Fig.~\ref{fig:rd_curves} presents the RD curves of our proposed 4D-MoDe compared with 4DGC~\cite{4DGC}, TeTriRF~\cite{tetrirf}, ReRF~\cite{rerf}, and 3DGStream~\cite{3dgstream} on representative sequences including \textit{Coffee Martini}, \textit{Flame Steak}, \textit{Alexa}, and \textit{Trimming}. 4D-MoDe demonstrates superior RD performance, particularly in the low bitrate regime. Notably, it remains effective even at extremely low bitrates where other methods either fail to operate or experience severe quality degradation. Moreover, 4D-MoDe exhibits the steepest RD curve slope, indicating the most rapid quality improvement with increasing bitrate, which further highlights its outstanding RD efficiency. To quantitatively validate this advantage, we compute the BDBR relative to TeTriRF~\cite{tetrirf} and 4DGC~\cite{4DGC}, as summarized in Tab.~\ref{table:BDBR}. Compared to TeTriRF, 4D-MoDe achieves an average BDBR reduction of \textbf{94.45\%} on the N3DV dataset. On the MeetRoom and Google Immersive datasets, it yields BDBR savings of \textbf{91.49\%} and \textbf{95.70\%}, respectively. Similarly, significant bitrate savings are observed when compared to 4DGC, consistently confirming the superior RD performance of our approach across diverse scenarios.

\begin{table}[tb]
\centering
\caption{The BDBR and BD-PSNR results of our 4D-MoDe and 4DGC\cite{4DGC} when compared with TeTriRF \cite{tetrirf} on three datasets}
\label{table:BDBR}
\scalebox{1}{
\begin{tabular}{c|c|cc}
\hline
Dataset                   & Method & BDBR(\%)$\downarrow$   & BD-PSNR(dB)$\uparrow$ \\ \hline
\multirow{2}{*}{N3DV \cite{n3dv}}     
                          & 4DGC    & -77.70 & 0.971 \\
                          & Ours    & \textbf{-94.452} & \textbf{5.372} \\ \hline
\multirow{2}{*}{MeetRoom \cite{StreamRF}} 
                          & 4DGC    & -40.71 & 0.55 \\
                          & Ours    & \textbf{-91.492} & \textbf{1.652} \\ \hline
\multirow{2}{*}{\begin{tabular}[c]{@{}c@{}}Google Immersive \cite{googleimmersive}\end{tabular}}   
                          & 4DGC    & -59.99 & 1.03 \\
                          & Ours    & \textbf{-95.70} & \textbf{2.426} \\ \hline
\end{tabular}
}
\end{table}

Tab.~\ref{tab:time} compares the computational efficiency of 4D-MoDe with state-of-the-art dynamic scene compression methods, ReRF~\cite{rerf} and TeTriRF~\cite{tetrirf}, across training, rendering, encoding, and decoding stages. 4D-MoDe achieves a training time of 0.68 minutes, which is 63$\times$ and 1.5$\times$ faster than ReRF and TeTriRF, respectively. The rendering time is 0.006 seconds per frame, yielding 83$\times$ and 62$\times$ speedups over ReRF and TeTriRF, respectively. Furthermore, 4D-MoDe shows the shortest encoding (0.72 s) and decoding (0.09 s) times, indicating its potential for low-latency streaming applications.

\begin{table}[tb]
\centering
\caption{Complexity comparison of our 4D-MoDe method with the state-of-the-art dynamic scene compression methods, ReRF \cite{rerf} and TeTriRF \cite{tetrirf}.}
\label{tab:time}
\scalebox{1}{
\begin{tabular}{c|cccc}
\hline
Method    & Train(min) & Render(s) & Encode(s) & Decode(s) \\ \hline
ReRF \cite{rerf}    & 42.73      & 0.502     & 3.03      & 0.28      \\
TeTriRF \cite{tetrirf} & 1.04       & 0.375     & 0.79      & 0.31      \\
Ours    & \textbf{0.68}       & \textbf{0.006}     & \textbf{0.72}      & \textbf{0.09}      \\ \hline
\end{tabular}
}
\end{table}

\begin{figure*}[htb]
    \centering
    \includegraphics[width=\linewidth]{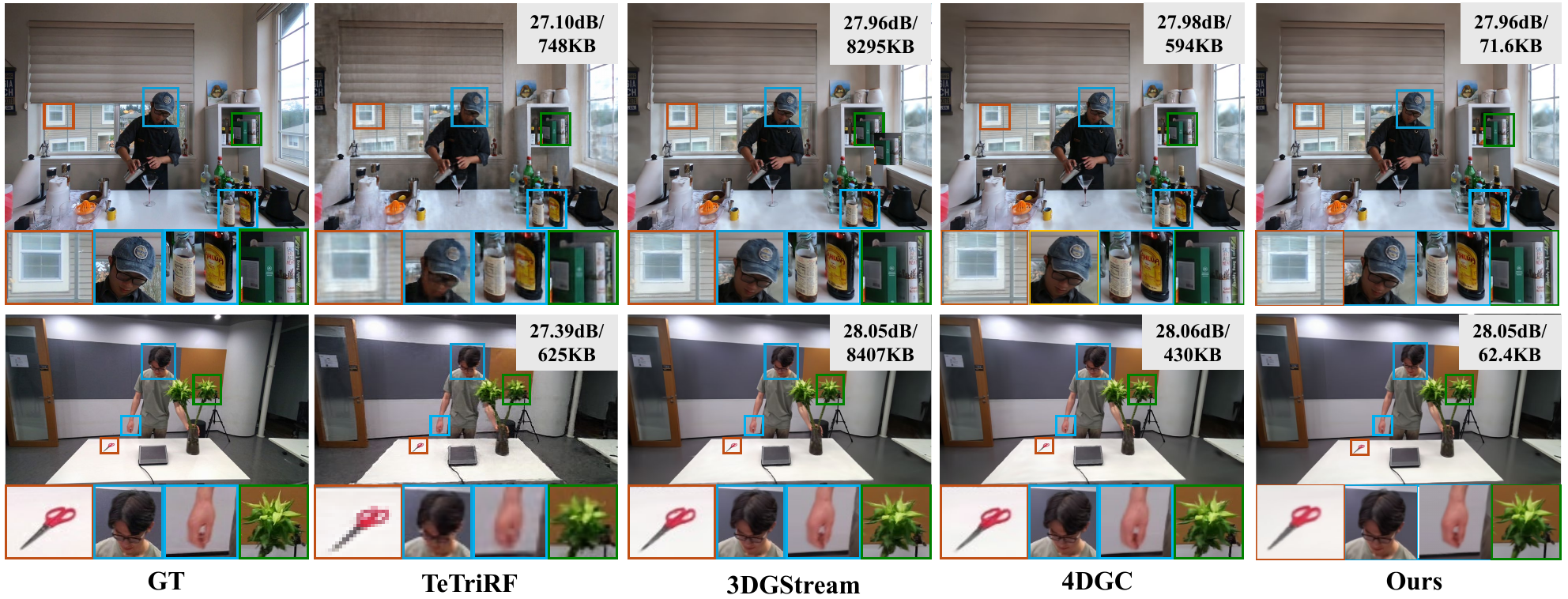}
    \caption{Qualitative comparison on the N3DV\cite{n3dv} and MeetRoom\cite{StreamRF} datasets against ReRF\cite{rerf}, TeTriRF\cite{tetrirf} and 3DGStream\cite{3dgstream}}
    \label{fig:compare_quality}
\end{figure*}

\textbf{Qualitative comparisons.}
We present a qualitative comparison on the \textit{coffee\_martini} and \textit{trimming} sequences against TeTriRF~\cite{tetrirf}, 3DGStream~\cite{3dgstream}, and 4DGC~\cite{4DGC}, as shown in Fig.\ref{fig:compare_quality}. Our method, 4D-MoDe, achieves comparable reconstruction quality to 3DGStream~\cite{3dgstream} and 4DGC~\cite{4DGC} while operating at an extremely low bitrate of only 60–70KB per frame, significantly outperforming others in compression efficiency.  Compared to TeTriRF~\cite{tetrirf},  4D-MoDe more effectively preserves fine details in both sequences. In \textit{coffee\_martini}, textures such as window blinds, facial features, and bottle labels remain sharp, while TeTriRF~\cite{tetrirf} shows noticeable blurring and smoothing. In \textit{trimming}, our method retains clear edges on the scissors, face, and plant, whereas TeTriRF~\cite{tetrirf} suffers from detail loss and visual artifacts. These results demonstrate that 4D-MoDe maintains excellent spatial fidelity and temporal coherence under extreme compression, offering over \textbf{7$\times$} bitrate reduction compared to 4DGC~\cite{4DGC} while retaining perceptual quality.

\subsection{Evaluation}
\textbf{Ablation Studies.} To assess the individual contributions of key components in our framework, we conduct five ablation studies by selectively disabling motion decoupling, motion estimation, Gaussian compensation, adaptive grouping, and entropy-guided joint optimization during training.
(1) Motion Decoupling: We disable the motion decoupling module and treat all Gaussians in the keyframe as dynamic, foregoing the separation between static backgrounds and dynamic foregrounds.  
(2) Motion Estimation: We remove the motion estimation module entirely, and rely solely on per-frame optimized compensated Gaussians without inter-frame tracking. 
(3) Gaussian Compensation: We retain motion estimation but disable the compensation mechanism that supplements Gaussians in regions with high motion residuals.  
(4) Adaptive Grouping: We replace the adaptive grouping strategy with a fixed GOP size of 50, thereby preventing content-aware keyframe updates.  
(5) Joint Optimization: We remove entropy-guided supervision by decoupling the training of the motion representation and the entropy model, which are optimized sequentially rather than jointly.

\begin{figure}[htb]
    \centering
    \includegraphics[width=\linewidth]{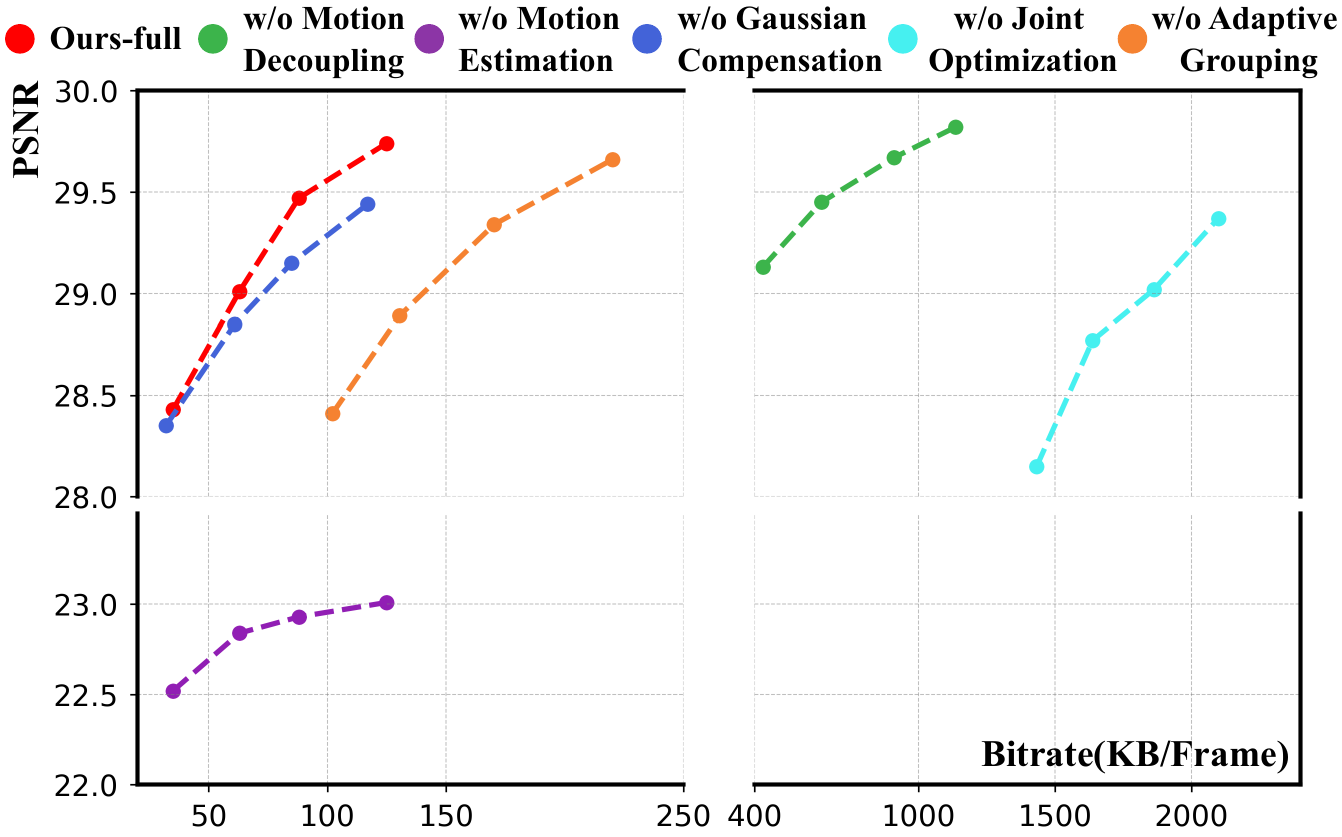}
    \caption{Ablation study on the impact of motion decoupling, motion estimation, Gaussian compensation, adaptive grouping, and entropy-guided joint optimization.}
    \label{fig:ablation}
\end{figure}
\begin{figure}[tb]
    \centering
    \includegraphics[width=1\linewidth]{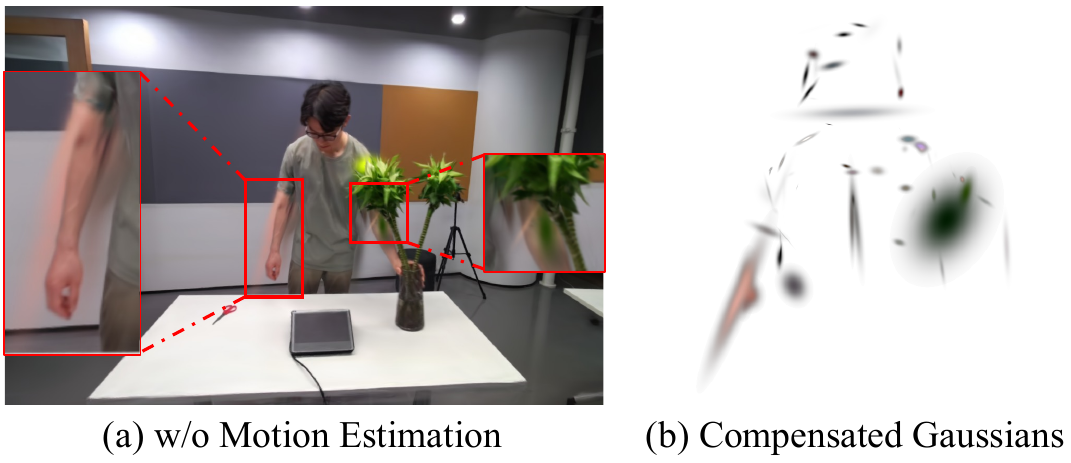}
    \caption{Qualitative illustration of the importance of motion estimation. Without it, dynamic Gaussians from previous frames accumulate and interfere with compensated Gaussians, resulting in visible ghosting.}
    \label{fig:me_ablation}
\end{figure}

As illustrated in Fig.~\ref{fig:ablation}, all variants exhibit degraded RD performance to varying degrees, confirming the necessity of each component. Disabling Gaussian compensation (\textcolor[HTML]{4363D8}{blue curve}) results in a consistent PSNR drop across all bitrate levels, particularly harming fine-grained motion regions. Removing motion estimation (\textcolor[HTML]{3CB44B}{green curve}) causes the most severe degradation, as outdated Gaussians from previous frames cannot be updated in time, leading to spatial overlaps and temporal artifacts. This phenomenon is further demonstrated in Fig.~\ref{fig:me_ablation}, where the lack of motion estimation results in visible ghosting due to overlap between stale and newly compensated Gaussians. Disabling motion decoupling (\textcolor[HTML]{911EB4}{purple curve}) increases model complexity, as all Gaussians are treated as dynamic, thereby enlarging the motion field and compensation overhead. Removing adaptive grouping (\textcolor[HTML]{F58231}{orange curve}) by fixing the GoP size to a static interval degrades performance due to the loss of content-aware keyframe insertion, which reduces temporal adaptability and affects RD efficiency. Lastly, removing entropy-guided supervision (\textcolor[HTML]{46F0F0}{cyan curve}) significantly increases the bitrate, highlighting its importance in driving the motion field towards a compact and compressible representation.

Tab.~\ref{tab:abla_bdbr} quantifies the impact of each component using BDBR and BD-PSNR metrics on the Google Immersive dataset. Removing motion estimation results in a BD-PSNR drop of \textbf{6.256 dB}, the most severe among all variants. Gaussian compensation contributes a \textbf{14.1\%} reduction in bitrate, while joint optimization of entropy and motion representation improves compression by preventing redundancy in the learned grid. The entry marked ‘--’ denotes failure to compute the metric due to unaligned RD curves.

\begin{table}[tb]
\setlength{\tabcolsep}{1.5pt} 
\centering
\caption{BDBR and BD-PSNR for ablation variants on Google Immersive dataset.}
\label{tab:abla_bdbr}
\scriptsize 
\begin{tabular}{c|c|c|c|c|c}
\hline
Metric & w/o Decouple & w/o Estim. & w/o Comp. & w/o Group. & w/o JointOpt \\
\hline
BDBR (\%)    & 82.47   & --      & 14.10   & 18.66   & 2982.77 \\
BD-PSNR (dB) & -0.51   & -6.26   & -0.14   & -0.23   & -189.64 \\
\hline
\end{tabular}
\end{table}

\begin{figure}[tb]
    \centering
    \includegraphics[width=\linewidth]{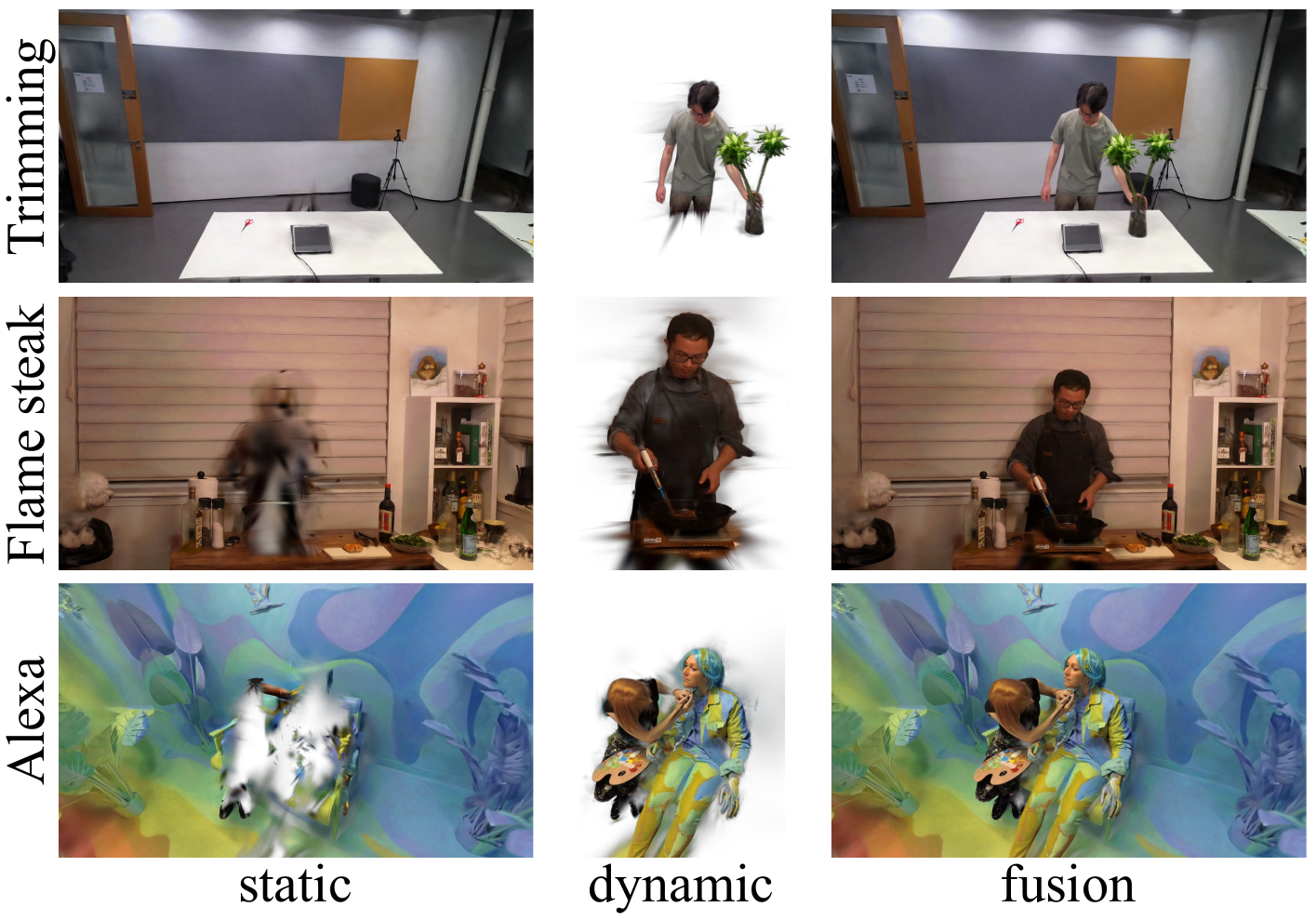}
    \caption{Visualization of our static and dynamic point separation on Trimming, Flame Steak and Alexa scene}
    \label{fig:DecompViz}
\end{figure}

\textbf{Motion Decoupling.} As illustrated in Fig.~\ref{fig:DecompViz}, we visualize static and dynamic Gaussians independently on the \textit{Flame\_steak} and \textit{Alexa} sequences to assess the effectiveness of our motion decoupling strategy. The separated renderings show that 4D-MoDe accurately disentangles moving foregrounds from static backgrounds in a fully unsupervised manner. In \textit{Flame\_steak}, dynamic components such as the chef and cooking flames are clearly isolated, while static elements like spice jars and shelves remain stable. Similarly, in \textit{Alexa}, human interactions are captured as dynamic content, with the stylized background preserved as static. The fused renderings validate the compositional integrity of the decomposition, maintaining structural consistency and temporal coherence. These results demonstrate that 4D-MoDe enables robust content-level separation, supporting scalable compression, foreground-only streaming, and flexible scene editing.

\begin{figure}[tb]
    \centering
    \includegraphics[width=0.9\linewidth]{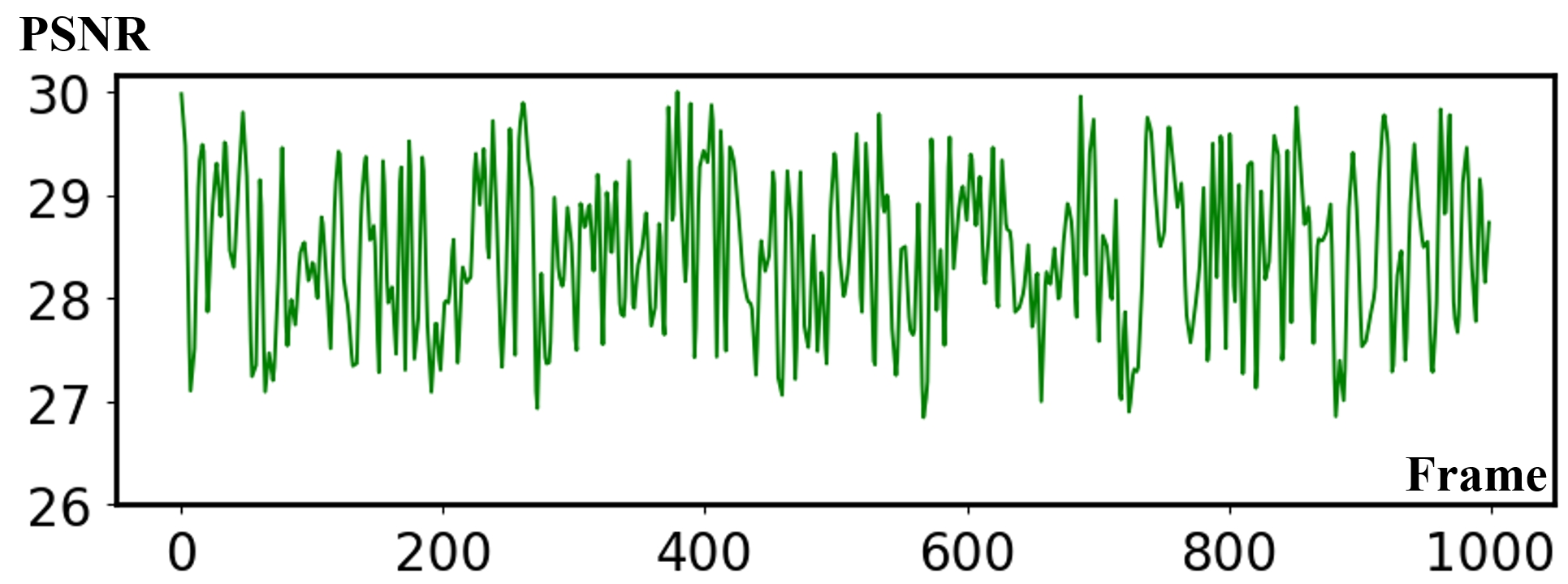}
    \caption{Frame-wise PSNR performance on a 1000-frame \textit{Flame Salmon} sequence.}
    \label{fig:longsalmon}
\end{figure}

\textbf{Long Sequence.} To evaluate the temporal stability of 4D-MoDe, we present the per-frame PSNR on a challenging 1000-frame dynamic sequence (\textit{Flame Salmon}), as shown in Fig.~\ref{fig:longsalmon}. Despite the long duration and complex motion patterns, 4D-MoDe consistently maintains high-quality reconstruction with minimal fluctuations. This indicates that our motion-decoupled modeling and compensation framework generalizes well over extended sequences.

\textbf{Storage Breakdown.} Tab.~\ref{tab:bitstream-breakdown} presents the size breakdown of our 4D-MoDe encoded bitstream on the N3DV dataset, categorized into keyframe and inter-frame components. Within each GoP, the compressed static Gaussian background in the keyframe occupies  6348 KB, while the dynamic foreground in the keyframe requires only 686~KB. For each inter-frame, the compensated Gaussians account for \textbf{2.5~KB}, and the associated motion information occupies \textbf{8.9~KB}. These results highlight the compactness of inter-frame data achieved through our motion decoupling strategy, which effectively reduces temporal redundancy and inter-frame bitrate, thereby facilitating efficient streaming with minimal overhead.

\begin{table}[htb]
\setlength{\tabcolsep}{2pt} 
\centering
\caption{Bitstream size breakdown for keyframe and inter-frame components on N3DV dataset.}
\begin{tabular}{c|c|c|c}
\hline
\begin{tabular}[c]{@{}c@{}}Static Gaussians \\ (Keyframe)\end{tabular} & \begin{tabular}[c]{@{}c@{}}Dynamic Gaussians\\ (Keyframe)\end{tabular} & \begin{tabular}[c]{@{}c@{}}Compensation\\ (Inter-Frames)\end{tabular} & \begin{tabular}[c]{@{}c@{}}Motion \\ (Inter-Frames)\end{tabular} \\ \hline
6348KB & 686 KB & 2.5 KB & 8.9 KB \\ \hline
\end{tabular}
\label{tab:bitstream-breakdown}
\end{table}

\section{Conclusion}
This paper presents 4D-MoDe, a motion-decoupled 4D Gaussian compression framework designed for scalable and editable volumetric video systems. By explicitly separating dynamic foregrounds from static backgrounds, our approach achieves high-quality rendering while significantly improving compression efficiency and editability. The proposed framework introduces systematic innovations in scene representation, motion modeling, end-to-end joint optimization, and streamable compression strategy, establishing a compact yet interactive pipeline for dynamic volumetric content. Extensive experiments demonstrate that 4D-MoDe achieves superior compression rates and reconstruction quality across diverse real-world scenes. Compared to existing methods, our approach maintains high visual fidelity while reducing transmission bandwidth. Furthermore, the motion-decoupled representation enables a range of new applications, including background replacement, foreground-only streaming, and real-time interactive rendering, highlighting the practical potential of our method in real-world volumetric video systems.

\bibliographystyle{IEEEtran}
\bibliography{reference}

\vspace{-2cm}
\begin{IEEEbiography}[
{\includegraphics[width=1in,height=1.25in,clip,keepaspectratio]{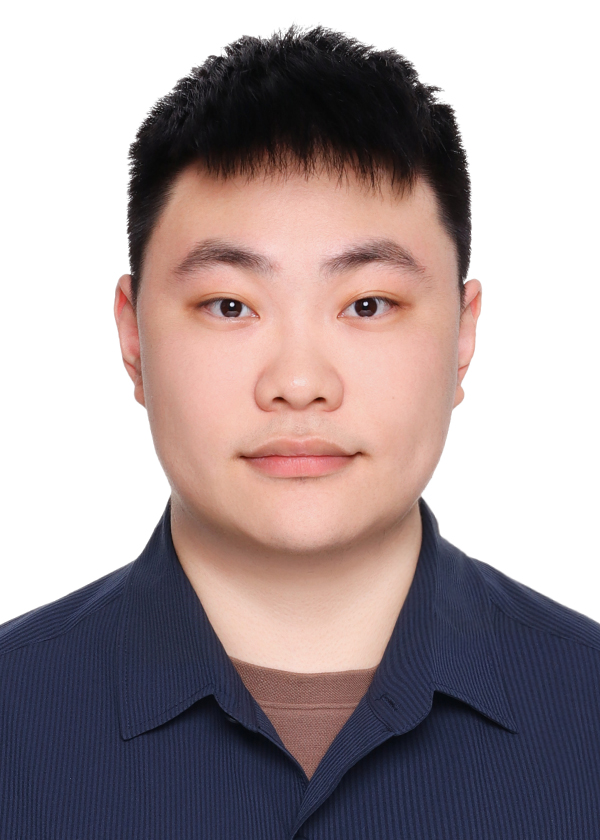}}
]{Houqiang Zhong} (Graduate Student Member, IEEE) received his B.E. degree from Shanghai University in 2020 and M.S. degree from ShanghaiTech University in 2023. He is currently pursuing a Ph.D. in the School of Electronic Information and Electrical Engineering at Shanghai Jiao Tong University. His research focuses on volumetric video and 3D reconstruction.
\end{IEEEbiography}

\begin{IEEEbiography}[
{\includegraphics[width=1in,height=1.25in,clip,keepaspectratio]{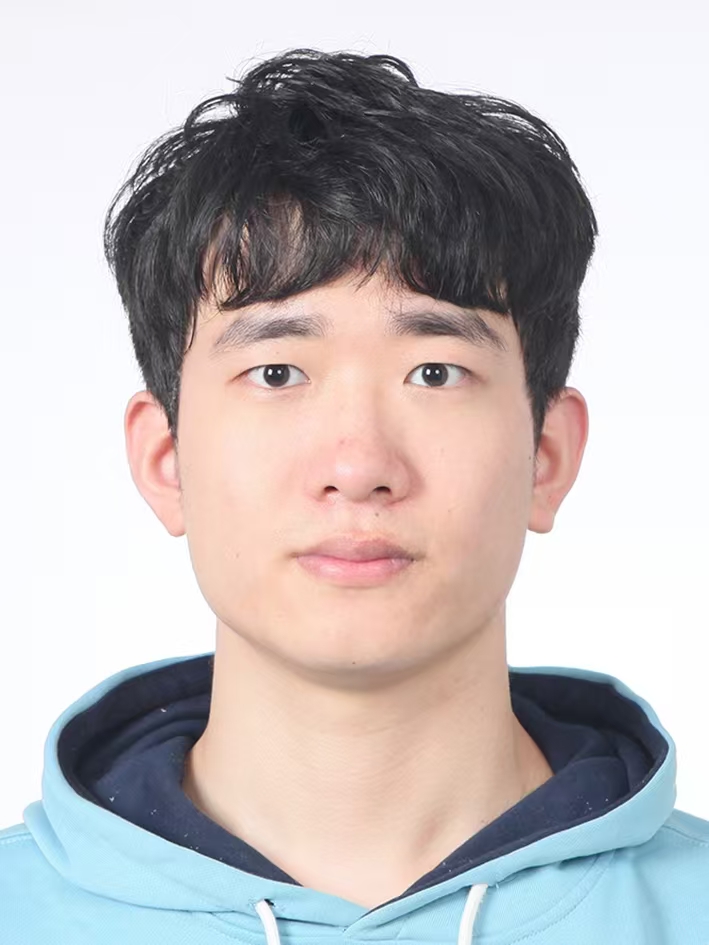}}
]{Zihan Zheng} (Student Member, IEEE) received his B.E. degree from Shanghai Jiao Tong University in 2020. He is currently pursuing a Ph.D. in the School of Electronic Information and Electrical Engineering at Shanghai Jiao Tong University. His research focuses on volumetric video, 3D reconstruction and 3D generation.
\end{IEEEbiography}

\begin{IEEEbiography}[
{\includegraphics[width=1in,height=1.25in,clip,keepaspectratio]{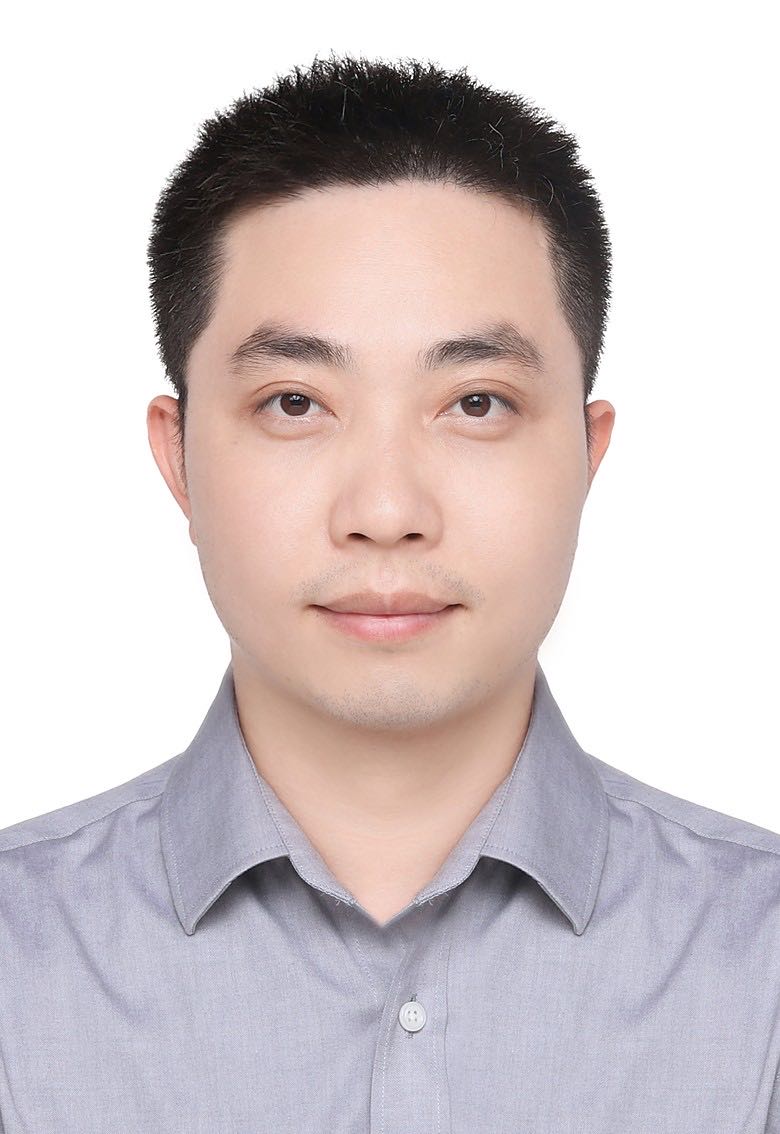}}
]{Qiang Hu} (Member, IEEE) received the B.S. degree in electronic engineering from University of Electronic Science and Technology of China in 2013, and the Ph.D. degree in information and communication engineering from Shanghai Jiao Tong University in 2019.  He is currently an Assistant Researcher at Cooperative Medianet Innovation Center, Shanghai Jiao Tong University. Before that, he was an Assistant Researcher at ShanghaiTech University from 2021 to 2023. He was a Postdoc Researcher at ShanghaiTech University from 2019 to 2021. His research interests focus on 2D/3D video compression, 3D reconstruction, and generative intelligence media. His works have been published in top-tier journals and conferences, such as IEEE Transactions on Image Processing (TIP), Conference on Computer Vision and Pattern Recognition (CVPR).
\end{IEEEbiography}

\begin{IEEEbiography}[
{\includegraphics[width=1in,height=1.25in,clip,keepaspectratio]{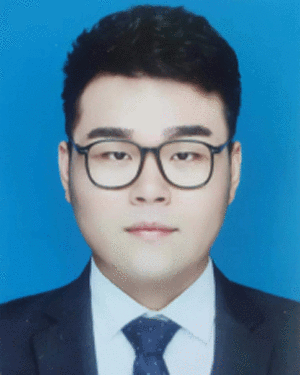}}
]{Yuan Tian} received the B.Sc. degree in electronic engineering from Wuhan University, Wuhan, China, in 2017. He obtained the Ph.D. degree with the Department of Electronic Engineering, Shanghai Jiao Tong University, Shanghai, China, in 2023. Currently, he is a researcher at Shanghai Artificial Intelligence Laboratory. His works have been published in top-tier journals and conferences (e.g., TPAMI, IJCV, TIP, CVPR, ICCV, ECCV and AAAI). His research interests include video processing and generative AI.
\end{IEEEbiography}

\begin{IEEEbiography}[{\includegraphics[width=1in,height=1.25in,clip,keepaspectratio]{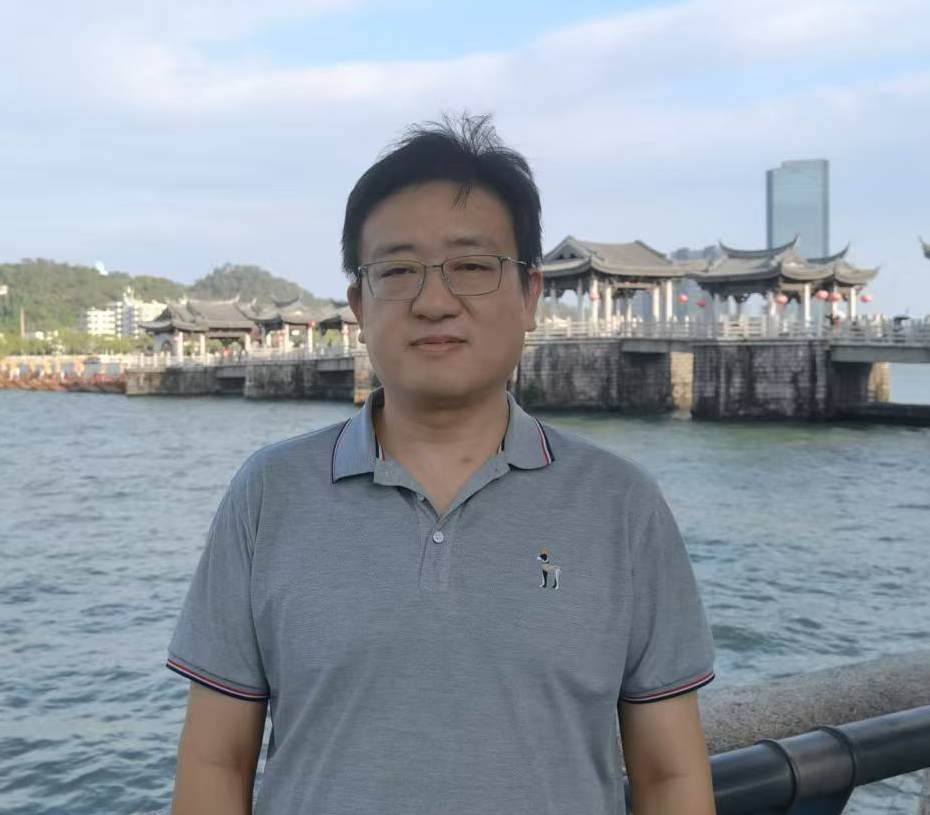}}]{Ning Cao } is employed at E-surfing Vision Technology Co., Ltd (a subsidiary of China Telecom). His standardization contributions span ITU-T, CCSA, and TC100, where he notably led the development of ITU-T F.743 and F.743.1 standards. He holds five provincial/ministerial-level awards and over 20 international/domestic patents in related fields.
\end{IEEEbiography}

\begin{IEEEbiography}[{\includegraphics[width=1in,height=1.25in,clip,keepaspectratio]{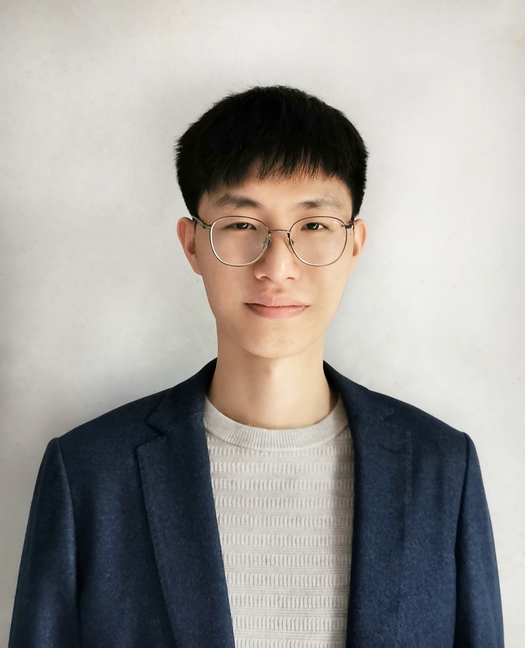}}]{Lan Xu} received the B.E. degree from Zhejiang University in 2011 and the Ph.D. degree from the Department of Electronic and Computer Engineering (ECE), The Hong Kong University of Science and Technology (HKUST), in 2020. He is currently an Assistant Professor with ShanghaiTech University. His research interests include computer vision, computer graphics, and machine learning.
\end{IEEEbiography}

\begin{IEEEbiography}[
{\includegraphics[width=1in,height=1.25in,clip,keepaspectratio]{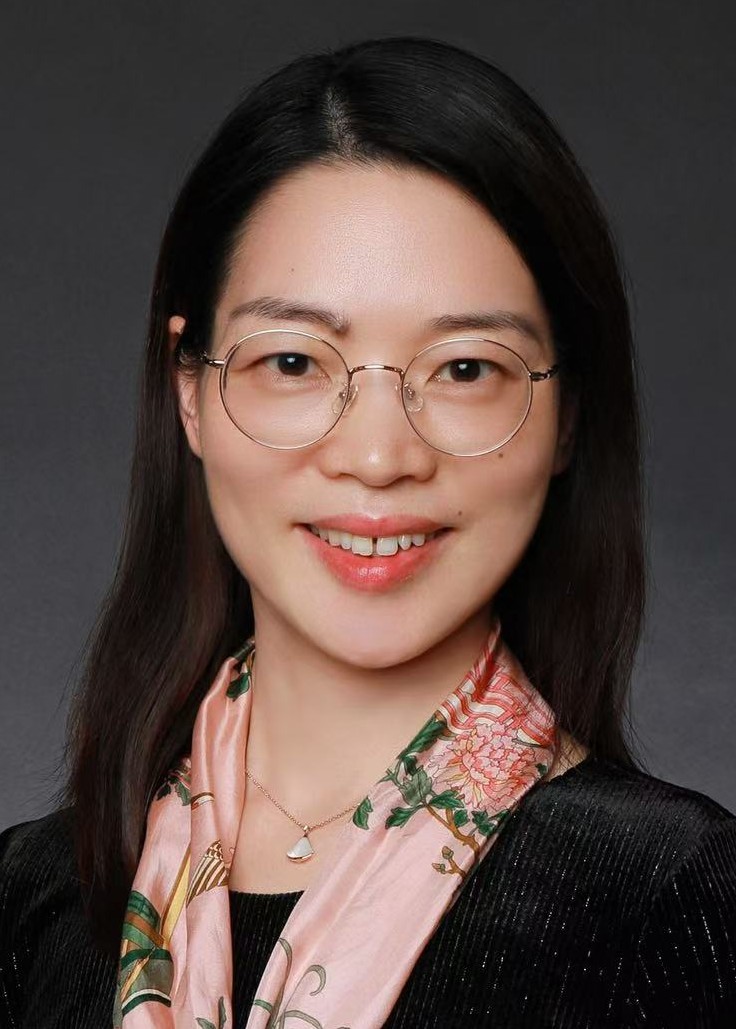}}
]{xiaoyun Zhang}( Member, IEEE), Professor of Cooperative Medianet Innovation Center (CMIC), Shanghai Jiao Tong University. She received her Ph.D (supervised by Yuncai Liu) on Pattern Recognition from Shanghai Jiao Tong University and Master (supervised by Zongben Xu) on Applied Mathematics from Xi`an Jiao Tong University respectively. She has co-authorized papers on IEEE TPAMI, TIP, CVPR, ICCV, etc., and her Ph.D. thesis has been nominated as “National 100 Best Ph.D. Theses of China”. She has been a visiting scholar of Harvard University for one year. She is also a member of State Key Laboratory of UHD video and audio production and presentation, and has collaboration with CCTV in video restoration and enhancement.
\end{IEEEbiography}

\begin{IEEEbiography}[{\includegraphics[width=1in,height=1.25in,clip,keepaspectratio]{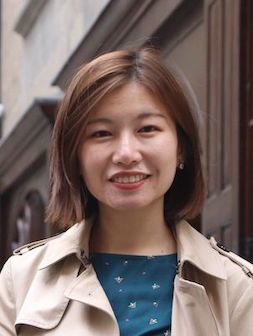}}]{Zhengxue Cheng}
receives the B.E. degree from Shanghai Jiao Tong University, Shanghai, China in 2014 and the M.E. degrees from Waseda University, Kitakyushu, Japan and Shanghai Jiao Tong University in 2015 and 2017, respectively through a double-degree program. She receives a PhD.degree at Waseda University, Tokyo, Japan in 2020. Then She works in Ant Group, Hangzhou, China, as an Algorithm Expert until April 2024. She joined the institute of Image Communication and Network Engineering, Shanghai Jiao Tong University as an assistant researcher in May 2024. Her research interests include deep learning-based media compression and quality evaluation.
\end{IEEEbiography}

\begin{IEEEbiography}
[{\includegraphics[width=1in,height=1.25in,clip,keepaspectratio]{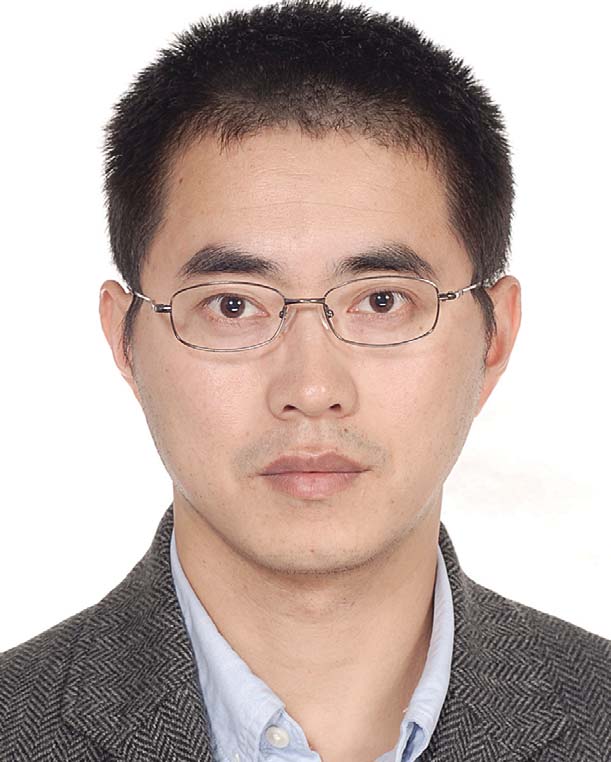}}]{Li Song}
(Senior Member, IEEE) received B.E. and M.S. degrees in engineering in 1997 and 2000, respectively, and a Ph.D. degree in electrical engineering from Shanghai Jiao Tong University in 2005, where he is currently a Full Professor with the Department of Electronic Engineering.
He was also a Visiting Professor at Santa Clara University from 2011 to 2012.
He has more than 200 publications, more than 50 granted patents, and 18 standard technical proposals in the field of video coding and image processing.
His research interests include video processing and multimedia systems.
He was a recipient of the National Science and Technology Progress Award in 2015, the Okawa Foundation Research Grant in 2012, the Second Place Award from IEEE ICME-Twitch grand challenge in 2017, the Best 10\% Paper Award from IEEE VCIP in 2016, and the Best Paper Award from the International Conference on Wireless Communications and Signal Processing in 2010.
He has served as an Associate Editor for Multidimensional Systems and Signal Processing since 2012 and a Guest Editor for IEEE Transaction on Broadcasting, a special issue on the quality of experience of advanced broadcast services, in June 2018.
He has served as the area or session chair for various international conferences and workshops.
\end{IEEEbiography}

\begin{IEEEbiography}[{\includegraphics[width=1in,height=1.25in,clip,keepaspectratio]{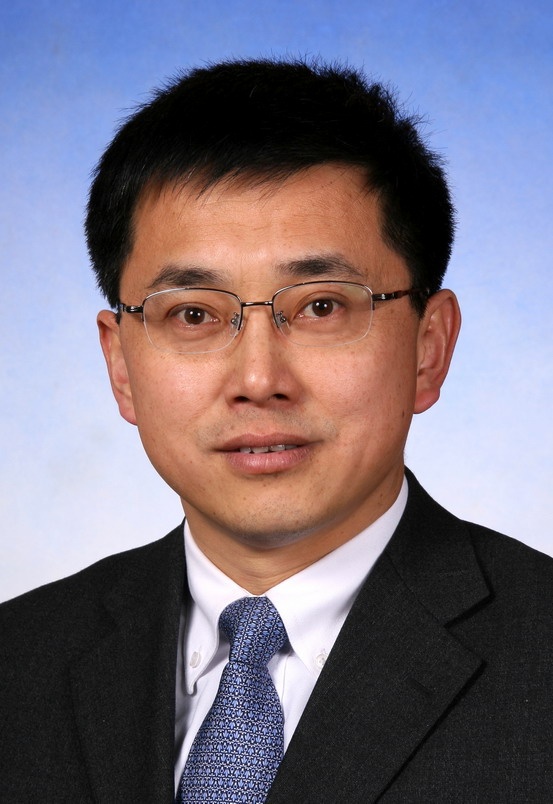}}]{Wenjun Zhang}
(Fellow, IEEE) received B.S., M.S., and Ph.D. degrees in electronic engineering from Shanghai Jiao Tong University, Shanghai, China, in 1984, 1987, and 1989, respectively.
From 1990 to 1993, he worked as a Postdoctoral Fellow with Philips, Nuremberg, Germany, where he was actively involved in developing the HD-MAC system.
He joined the faculty of Shanghai Jiao Tong University in 1993 and became a Full Professor of Electronic Engineering in 1995. He is the Chief Scientist of the Chinese Digital TV Engineering Research Centre, an industry/government consortium in DTV technology research and standardization.
His main research interests include digital video coding and transmission, multimedia semantic processing, and intelligent video surveillance.
\end{IEEEbiography}

\end{document}